\pdfoutput=1
\documentclass[11pt]{article}

\usepackage[]{acl}

\usepackage{times}
\usepackage{latexsym}

\usepackage[T1]{fontenc}
\usepackage[utf8]{inputenc}

\usepackage{microtype}

\usepackage{inconsolata}

\usepackage{mathtools}
\usepackage{times}
\usepackage{latexsym}

\usepackage{import}
\usepackage{hyperref}
\usepackage{graphicx}
\usepackage{adjustbox}
\usepackage{multirow}
\usepackage{verbatim}
\usepackage{algorithmic}
\usepackage{algorithm}
\usepackage{balance}
\usepackage{rotating}
\usepackage{xcolor}
\usepackage{enumitem}
\usepackage{footmisc}
\usepackage{caption}
\usepackage{subcaption}
\usepackage{colortbl}

\title{Redundancy Aware Multiple Reference Based\\ Gainwise Evaluation of Extractive Summarization}

\author{Mousumi Akter \\
  Research Center Trustworthy\\
Data Science and Security \\
  Technical University Dortmund, Germany \\
  \texttt{mousumi.akter@tu-dortmund.de} \\\And
  Santu Karmaker \\
  Big Data Intelligence (BDI) Lab\\ Auburn University\\ Alabama, USA\\
  \texttt{sks0086@auburn.edu} \\}

\begin{document}
\maketitle

\begin{abstract}
The ROUGE metric is commonly used to evaluate extractive summarization task, but it has been criticized for its lack of semantic awareness and its ignorance about the ranking quality of the extractive summarizer. Previous research has introduced a gain-based automated metric called \textit{Sem-nCG} that addresses these issues, as it is both rank and semantic aware. However, it does not consider the amount of redundancy present in a model summary and currently does not support evaluation with multiple reference summaries. It is essential to have a model summary that balances importance and diversity, but finding a metric that captures both of these aspects is challenging. In this paper, we propose a redundancy-aware \textit{Sem-nCG} metric and demonstrate how the revised \textit{Sem-nCG} metric can be used to evaluate model summaries against multiple references as well which was missing in previous research. Experimental results demonstrate that the revised \textit{Sem-nCG }metric has a stronger correlation with human judgments compared to the previous \textit{Sem-nCG} metric and traditional ROUGE and BERTScore metric for both single and multiple reference scenarios.

\end{abstract}

\section{Introduction}
For the past two decades, ROUGE~\cite{lin-2004-rouge} has been the most used metric for evaluating extractive summarization tasks. Nonetheless, ROUGE has long been criticized for its lack of semantic awareness~\cite{DBLP:conf/emnlp/Graham15,DBLP:conf/emnlp/NgA15,DBLP:journals/corr/abs-1803-01937,DBLP:conf/acl/YangLLLL18} and its ignorance about the ranking quality of the extractive summarizer~\cite{DBLP:conf/acl/AkterBS22}.

To address these issues, previous work has proposed a gain-based metric called \textit{Sem-nCG}~\cite{DBLP:conf/acl/AkterBS22} to evaluate extractive summaries by incorporating rank and semantic awareness. Redundancy, a crucial factor in evaluating extractive summaries, was not, however, included in the \textit{Sem-nCG} metric. Additionally, their proposed \textit{Sem-nCG} metric does not support the evaluation of model summaries against multiple references. However, it is well recognized that a set of documents can have multiple, very different, and equally valid summaries; as such, obtaining multiple reference summaries can improve the stability of the evaluation~\cite{DBLP:conf/aaai/Nenkova05, DBLP:conf/ntcir/Lin04}. It's quite challenging to come up with a metric that takes into account the balance between importance and diversity in model summary. Therefore, it's necessary to carry out a systematic study on how to integrate redundancy and multiple references to the existing \textit{Sem-nCG} metric.

In this paper, we first incorporate redundancy into the previously proposed \textit{Sem-nCG} metric. In other words, we propose a redundancy-aware \textit{Sem-nCG} metric by exploring different ways of incorporating redundancy into the original metric. Through extensive experiments, we demonstrate that the redundancy-aware \textit{Sem-nCG} exhibits a notably stronger correlation with humans than the original \textit{Sem-nCG} metric. 

Next, we demonstrate how this redundancy-aware metric could be applied to evaluate model summaries against multiple references. This is a non-trivial task because \textit{Sem-nCG} evaluates a model-generated summary by considering it as a ranked list of sentences and then comparing it against an automatically inferred \textit{ground-truth} ranked list of sentences within a source document based on a single human written summary~\cite{DBLP:conf/acl/AkterBS22}. However, in the case of multiple references, the \textit{ground-truth} ranked list of source sentences must be inferred based on all available human-written reference summaries, not just one. 

When there are multiple reference summaries available, incorporating them into evaluation poses significant challenge. This is because the quality of human-written summaries differs not only in writing style but also in focus. Moreover, including multiple reference summaries with a lot of terminology variations and paraphrasing makes the automated evaluation metric less stable~\cite{DBLP:conf/lrec/CohanG16}. In this work, we have also shown how to infer a single/unique ground-truth ranking based on multiple reference summaries with the proposed redundancy-aware \textit{Sem-nCG} metric. Our findings suggest that, compared to the conventional ROUGE and BERTScore metric, the redundancy-aware \textit{Sem-nCG} exhibits a stronger correlation with human judgments for evaluating model summaries when both single and multiple references are available. Therefore, we encourage the community to use redundancy-aware \textit{Sem-nCG} to evaluate extractive summarization tasks. Our contributions are:

\begin{itemize}
    \item Redundancy of extracted sentences is a common problem in extractive summarization systems. We have demonstrated how to consider redundancy awareness in the already-designed \textit{Sem-nCG} metric.
    \item We present how to use the redundancy-aware \textit{Sem-nCG} metric for summary evaluation with multiple references which poses unique challenges of variability.
    \item The revised \textit{Sem-nCG} metric exhibits a stronger correlation with human judgments for evaluating model summaries when both single and multiple references are available, not only with the previous \textit{Sem-nCG} metric but also with conventional ROUGE and BERTScore metric.
\end{itemize}

\section{Redundancy-aware \textit{Sem-nCG} Metric} \label{method}
\noindent{\bf {\textit{Sem-nCG} Score:}} Normalized Cumulative Gain (\textit{nCG}) is a popular evaluation metric in information retrieval to evaluate the quality of a ranker. nCG compares the model ranking with an \textit{ideal} ranking and assigns a certain score to the model based on some pre-defined gain. \cite{DBLP:conf/acl/AkterBS22} has utilized the idea of \textit{nCG} in the evaluation of extractive summarization. The basic concept of Sem-nCG is to compute the gain (\textit{CG@k}) obtained by a top $k$ extracted sentences and divide that by the maximum/ideal possible gain (\textit{ICG@k}), where the gains are inferred by comparing the input document against a human written summary. Mathematically:

\vspace{-1 mm}
\begin{equation}\label{equ:nCG}\small
\textit{Sem-nCG@k}=\frac{\textit{CG@k}}{\textit{ICG@k}}
\end{equation} 

\noindent{\bf {Redundancy Score:}} We followed~\cite{DBLP:conf/acl/ChenLK20} to compute self-referenced redundancy score which is computationally efficient and less ambiguous than classical approaches. The summary, $X$, itself is used as the reference to determine the degree of semantic similarity between each summary token/sentence and the other tokens/sentences. The average of maximum semantic similarity is used to determine the redundancy score. For a given summary, $X = \{x_1, x_2,...,x_n\}$, the calculation is as follows:
\vspace{-1 mm}
\begin{equation}\label{equ:red}\small
Score\textsubscript{red}=\frac{\sum_{i}max_{j:i\neq j} Sim(x_{j}, x_{i})}{\textbf{|X|}}
\end{equation} 

where, $j:i\neq j$ denotes that the similarity between $x_{i}$ and itself has not been considered. Note that \textit{Score\textsubscript{red}} $\in [0, 1]$ in our case and lower is better.

\smallskip
\noindent{\bf {Final Score:}} We used the following formula to calculate the final score after obtaining the scores of \textit{Sem-nCG} and \textit{Score\textsubscript{red}}:
\vspace{-1 mm}
\begin{equation}\label{equ:final}\small
Score= \lambda*\textit{Sem-nCG} + (1 - \lambda)*(1 - \textit{Score\textsubscript{red}})
\end{equation} 

\noindent Here, $\lambda \in [0, 1]$ is a hyper-parameter to scale the weight between \textit{Score\textsubscript{red}} and \textit{Sem-nCG}. \textit{Score} $\in [0, 1]$ where higher score means better summary.

\section{Experimental Setup} \label{expt}
\noindent{\bf {Dataset:}} Human correlation is an essential attribute to consider while assessing the quality of a metric. To compute the human correlation of the revised redundancy-aware \textit{Sem-nCG} metric, we utilized SummEval dataset from~\cite{DBLP:journals/tacl/FabbriKMXSR21}\footnote{We used the dataset by~\cite{DBLP:journals/tacl/FabbriKMXSR21}, the only available benchmark "meta-evaluation dataset" for \textbf{extractive summarization}, to the best of our knowledge. \textit{Sem-nCG}'s authors have demonstrated its correlation with human judgment on this dataset. To ensure a fair comparison, we maintained the same settings as the original \textit{Sem-nCG} when assessing the redundancy-aware \textit{Sem-nCG}.}. The annotations include summaries generated by 16 models (abstractive and extractive) from 100 news articles (1600 examples in total) on the CNN/DailyMail Dataset. Each source news article includes the original CNN/DailyMail reference summary as well as 10 additional crowd-sourced reference summaries. Each summary was annotated by 5 independent crowd-sourced workers and 3 independent experts (8 annotations in total) along the four dimensions: \textit{Consistency}, \textit{Relevance}, \textit{Coherence} and \textit{Fluency}~\cite{DBLP:journals/tacl/FabbriKMXSR21}\footnote{See Appendix~\ref{hum_dimension} for details}. As this work focuses on the evaluation of extractive summarization, we considered the output generated by extractive models and filtered out samples comprising less than $3$ sentences (as we report \textit{Sem-nCG@3}). Additionally, we considered the expert annotations for the meta-evaluation, as non-expert annotations can be risky~\cite{gillick-liu-2010-non}. 

As was done in~\cite{DBLP:conf/acl/AkterBS22}, for each sample, from the 11 available reference summaries, we considered 3 settings: Less Overlapping Reference/LOR (highly abstractive references with fewer lexical overlap with the original document), Medium Overlapping Reference/MOR (medium lexical overlap with the original document) and Highly Overlapping Reference/HOR (highly extractive references with high lexical overlap with the original document).

\noindent{\bf {Embedding for Groundtruth Ranking:}} The core of the \textit{Sem-nCG} metric is to automatically create the groundtruth/ideal ranking against which the model ranking is compared. To create the groundtruth ranking, \cite{DBLP:conf/acl/AkterBS22} used various sentence embeddings. Similarly, we utilized various sentence embeddings as well since our goal is to compare the new redundancy-aware \textit{Sem-nCG} metric to the original \textit{Sem-nCG} metric. Specifically, we considered {Infersent} (v2)~\cite{DBLP:conf/emnlp/ConneauKSBB17}, {Semantic Textual Similarity benchmark (STSb - bert/roberta/distilbert)}~\cite{reimers-2019-sentence-bert}, {Elmo}~\cite{DBLP:conf/naacl/PetersNIGCLZ18} and {Google Universal Sentence Encoder (USE)}~\cite{cer-etal-2018-universal} with enc-2~\cite{iyyer-etal-2015-deep} based on the deep average network, to infer the groundtruth/ideal ranking of the sentences within the input document with guidance from the human written summaries.

\noindent{\bf {\textit{Score\textsubscript{red}} Computation:}} To compute the self-referenced redundancy score, we used the top-$3$ sentences from the model generated summary (as we report \textit{Sem-nCG@3}). We calculated each sentence's maximum similarity to other sentences and then averaged it to get the desired \textit{Score\textsubscript{red}}. We experimented with four distinct variations to compare the sentences: cosine similarity (by converting sentences to STSb-distilbert~\cite{reimers-2019-sentence-bert} embeddings), ROUGE~\cite{lin-2004-rouge}, MoverScore~\cite{DBLP:conf/emnlp/ZhaoPLGME19} and BERTScore~\cite{DBLP:conf/iclr/ZhangKWWA20}.
\begin{table*}[!htb]
\centering
\resizebox{\textwidth}{!}{
\begin{tabular}{ll|rrr|rrr|rrr|rrr}\hline
 &  & \multicolumn{3}{c|}{\textbf{Consistency}} & \multicolumn{3}{c|}{\textbf{Relevance}} & \multicolumn{3}{c|}{\textbf{Coherence}} & \multicolumn{3}{c}{\textbf{Fluency}} \\\cline{3-14}
\multirow{-2}{*}{\textbf{Embedding}} & \multirow{-2}{*}{\textbf{Type}} & \multicolumn{1}{l}{LOR} & \multicolumn{1}{l}{MOR} & \multicolumn{1}{l|}{HOR} & \multicolumn{1}{l}{LOR} & \multicolumn{1}{l}{MOR} & \multicolumn{1}{l|}{HOR} & \multicolumn{1}{l}{LOR} & \multicolumn{1}{l}{MOR} & \multicolumn{1}{l|}{HOR} & \multicolumn{1}{l}{LOR} & \multicolumn{1}{l}{MOR} & \multicolumn{1}{l}{HOR} \\ \hline
Inferesent & w/o redundancy & {0.08} & {0.06} & { 0.08} & {0.07} & {0.12} & {0.09} & {0.06} & {0.06} & {0.04} & {\cellcolor{green!15}{\textbf{0.05}}} & {0.03} & {\cellcolor{green!15}{\textbf{0.12}}} \\\hline
 & Cosine Similarity & 0.04 & 0.02 & 0.06 & 0.08 & 0.15 & 0.13 & 0.14 & 0.19 & 0.18 & 0.02 & -0.02 & 0.08 \\
 & ROUGE-1 & 0.07 & 0.05 & 0.11 & 0.11 & 0.18 & 0.17 & 0.18 & 0.25 & \cellcolor{green!15}{\textbf{0.26}} & -0.01 & -0.04 & 0.05 \\
 & MoverScore & 0.05 & 0.06 & 0.11 & 0.09 & 0.15 & 0.12 & 0.11 & 0.13 & 0.11 & 0.03 & 0.01 & 0.11 \\
\multirow{-4}{*}{\begin{tabular}[c]{@{}l@{}}+ Redundancy\\ penalty\end{tabular}} & BERTScore & 0.05 & 0.02 & 0.08 & 0.13 & 0.19 & 0.18 & 0.18 & 0.22 & 0.24 & -0.01 & -0.04 & 0.04 \\ \hline
Elmo & w/o redundancy & {0.06} & {0.07} & {0.09} & {0.02} & {0.08} & {0.06} & {0.02} & {0.02} & {0.01} & {0.00} & { 0.01} & {0.06} \\\hline
 & Cosine Similarity & 0.03 & 0.03 & 0.05 & 0.04 & 0.13 & 0.10 & 0.12 & 0.14 & 0.14 & -0.06 & -0.05 & 0.02 \\
 & ROUGE-1 & 0.08 & 0.05 & 0.08 & 0.07 & 0.15 & 0.14 & 0.17 & 0.20 & 0.20 & -0.06 & -0.06 & 0.01 \\
 & MoverScore & 0.08 & 0.07 & 0.10 & 0.04 & 0.10 & 0.09 & 0.07 & 0.06 & 0.06 & -0.02 & -0.01 & 0.05 \\
\multirow{-4}{*}{\begin{tabular}[c]{@{}l@{}}+ Redundancy\\ penalty\end{tabular}} & BERTScore & 0.06 & 0.03 & 0.05 & 0.09 & 0.17 & 0.16 & 0.17 & 0.19 & 0.18 & -0.06 & -0.07 & 0.00 \\\hline
STSb-bert & w/o redundancy & 0.11 & 0.08 & 0.09 & 0.03 & 0.13 & 0.12 & -0.01 & 0.06 & 0.01 & 0.03 & \cellcolor{green!15}{\textbf{0.10}} & 0.03 \\\hline
 & Cosine Similarity & 0.08 & 0.01 & 0.06 & 0.05 & 0.17 & 0.13 & 0.10 & 0.18 & 0.16 & -0.05 & 0.02 & 0.05 \\
 & ROUGE-1 & 0.12 & 0.05 & 0.09 & 0.08 & \cellcolor{green!15}{\textbf{0.22}} & 0.18 & 0.14 & 0.25 & 0.22 & -0.04 & -0.04 & 0.01 \\
 & MoverScore & 0.12 & 0.06 & 0.10 & 0.05 & 0.16 & 0.15 & 0.04 & 0.11 & 0.09 & -0.01 & 0.02 & 0.08 \\
\multirow{-4}{*}{\begin{tabular}[c]{@{}l@{}}+ Redundancy\\ penalty\end{tabular}} & BERTScore & 0.10 & 0.01 & 0.06 & 0.11 & \cellcolor{green!15}{\textbf{0.22}} & \cellcolor{green!15}{\textbf{0.20}} & 0.14 & 0.24 & 0.20 & -0.06 & -0.04 & 0.01 \\\hline
STSb-roberta & w/o redundancy & 0.12 & \cellcolor{green!15}{\textbf{0.14}} & 0.07 & 0.07 & 0.07 & 0.05 & 0.04 & 0.00 & -0.02 & -0.01 & 0.01 & 0.06 \\\hline
 & Cosine Similarity & 0.09 & 0.07 & { 0.05} & 0.08 & 0.11 & { 0.06} & 0.13 & 0.13 & 0.10 & -0.06 & -0.05 & { -0.01} \\
 & ROUGE-1 & 0.12 & 0.11 & 0.09 & 0.11 & 0.16 & 0.10 & 0.18 & 0.20 & 0.17 & -0.07 & -0.07 & -0.04 \\
 & MoverScore & 0.13 & 0.13 & 0.10 & 0.09 & 0.10 & 0.07 & 0.08 & 0.07 & 0.04 & -0.03 & 0.00 & 0.04 \\
\multirow{-4}{*}{\begin{tabular}[c]{@{}l@{}}+ Redundancy\\ penalty\end{tabular}} & BERTScore & 0.10 & 0.08 & 0.05 & 0.13 & 0.18 & 0.12 & 0.17 & 0.18 & 0.15 & -0.08 & -0.06 & -0.04 \\\hline
USE & w/o redundancy & 0.05 & 0.04 & 0.04 & 0.11 & 0.14 & 0.08 & 0.07 & 0.08 & 0.02 & 0.03 & 0.05 & 0.08 \\\hline
 & Cosine Similarity & 0.02 & -0.01 & 0.03 & 0.10 & 0.16 & 0.09 & 0.16 & 0.19 & 0.16 & -0.05 & 0.01 & 0.03 \\
 & ROUGE-1 & 0.06 & 0.02 & 0.07 & 0.13 & 0.21 & 0.14 & 0.20 & \cellcolor{green!15}{\textbf{0.26}} & 0.23 & -0.06 & 0.00 & 0.00 \\
 & MoverScore & 0.07 & 0.03 & 0.07 & 0.13 & 0.16 & 0.11 & 0.13 & 0.13 & 0.10 & 0.01 & 0.03 & 0.06 \\
\multirow{-4}{*}{\begin{tabular}[c]{@{}l@{}}+ Redundancy\\ penalty\end{tabular}} & BERTScore & 0.03 & -0.01 & 0.05 & \cellcolor{green!15}{\textbf{0.15}} & \cellcolor{green!15}{\textbf{0.22}} & 0.17 & \cellcolor{green!15}{\textbf{0.21}} & 0.24 & 0.22 & -0.06 & 0.00 & 0.00 \\ \hline
STSb-distilbert & w/o redundancy & 0.17 & 0.09 & \cellcolor{green!15}{\textbf{0.12}} & 0.06 & 0.09 & 0.07 & 0.06 & 0.03 & -0.01 & 0.01 & 0.03 & 0.04 \\\hline
 & Cosine Similarity & 0.16 & 0.04 & 0.06 & 0.07 & 0.12 & 0.07 & 0.14 & 0.16 & 0.11 & -0.05 & -0.03 & -0.04 \\
 & ROUGE-1 & 0.16 & 0.06 & 0.08 & 0.10 & 0.16 & 0.12 & 0.17 & 0.21 & 0.17 & -0.06 & -0.04 & -0.05 \\
 & MoverScore & \cellcolor{green!15}{\textbf{0.18}} & 0.08 & 0.10 & 0.08 & 0.12 & 0.09 & 0.09 & 0.09 & 0.04 & -0.02 & 0.01 & 0.01 \\
\multirow{-4}{*}{\begin{tabular}[c]{@{}l@{}}+ Redundancy\\ penalty\end{tabular}} & BERTScore & 0.14 & 0.03 & 0.05 & 0.12 & 0.18 & 0.14 & 0.17 & 0.20 & 0.16 & -0.06 & -0.05 & -0.05 \\\hline
Ensemble\textsubscript{sim} & w/o redundancy & 0.12 & 0.08 & 0.07 & 0.10 & 0.12 & 0.07 & 0.08 & 0.06 & 0.00 & 0.01 & 0.04 & 0.05 \\\hline
 & Cosine Similarity & 0.11 & 0.02 & 0.04 & 0.10 & 0.16 & 0.09 & 0.16 & 0.20 & 0.15 & -0.06 & -0.01 & -0.01 \\
 & ROUGE-1 & 0.13 & 0.05 & 0.08 & 0.13 & 0.21 & 0.14 & 0.20 & \cellcolor{green!15}{\textbf{0.26}} & 0.21 & -0.05 & -0.03 & -0.03 \\
 & MoverScore & 0.14 & 0.06 & 0.08 & 0.12 & 0.15 & 0.10 & 0.14 & 0.13 & 0.08 & -0.01 & 0.03 & 0.03 \\
\multirow{-4}{*}{\begin{tabular}[c]{@{}l@{}}+ Redundancy\\ penalty\end{tabular}} & BERTScore & 0.10 & 0.03 & 0.05 & \cellcolor{green!15}{\textbf{0.15}} & \cellcolor{green!15}{\textbf{0.22}} & 0.16 & \cellcolor{green!15}{\textbf{0.21}} & 0.25 & 0.20 & -0.05 & -0.02 & -0.03\\ \hline
\end{tabular}}
\caption{Kendall's tau ($\tau$) correlation coefficients of expert annotations for different embedding variations of \textit{Sem-nCG} along with various redundancy penalties when $\lambda = 0.5$. Low overlapping reference (LOR), medium overlapping CNN/DailyMail reference (MOR), and high overlapping reference (HOR) were chosen from 11 reference summaries per example to demonstrate the correlation. The highest value in each column is in bold green.}
\label{tab:hum}
\end{table*}

\begin{table*}[!htb]
\centering
\resizebox{\textwidth}{!}{
\begin{tabular}{l|rrr|rrr|rrr|rrr}
\hline
 & \multicolumn{3}{|c|}{Consistency} & \multicolumn{3}{c|}{Relevance} & \multicolumn{3}{c|}{Coherence} & \multicolumn{3}{c}{Fluency} \\\hline
\multirow{-2}{*}{} & \multicolumn{1}{|l}{LOR} & \multicolumn{1}{l}{MOR} & \multicolumn{1}{l|}{HOR} & \multicolumn{1}{l}{LOR} & \multicolumn{1}{l}{MOR} & \multicolumn{1}{l|}{HOR} & \multicolumn{1}{l}{LOR} & \multicolumn{1}{l}{MOR} & \multicolumn{1}{l|}{HOR} & \multicolumn{1}{l}{LOR} & \multicolumn{1}{l}{MOR} & \multicolumn{1}{l}{HOR} \\\hline
ROUGE-1 & {0.08} & {0.05} & {0.01} & {0.07} & {0.21} & {0.22} & {0.03} & {0.13} & {0.13} & {0.05} & {0.05} & {0.05} \\
ROUGE-L & 0.02 & 0.06 & -0.01 & 0.03 & 0.19 & 0.15 & -0.02 & 0.14 & 0.08 & 0.01 & 0.04 & -0.07 \\
BERTScore & 0.06 & 0.10 & 0.07 & 0.10 & 0.18 & 0.20 & 0.06 & 0.15 & 0.11 & 0.08 & 0.05 & 0.04 \\\hline
\end{tabular}
}
\caption{Kendall's tau correlation coefficients of ROUGE and BERTScore for Low overlapping reference (LOR), medium overlapping CNN/DailyMail reference (MOR), and high overlapping reference (HOR) chosen from 11 reference summaries per example to demonstrate the correlation.}
\label{tab:hum_rouge_bert}
\end{table*}

\setlength{\textfloatsep}{0pt}
\begin{figure*}[!htb]
        \begin{subfigure}[b]{\textwidth}
              \centering
             \frame{\includegraphics[width=\linewidth,trim={7 280 7 240},clip]{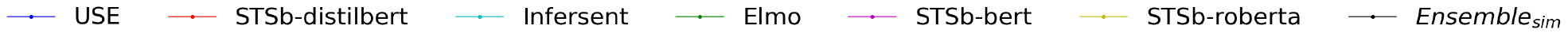}}
            \vspace{-4.5 mm}
        \end{subfigure}
        \hfill%
        \begin{subfigure}[b]{0.32\textwidth}
              \includegraphics[width=\linewidth,trim={10 10 155 105},clip]{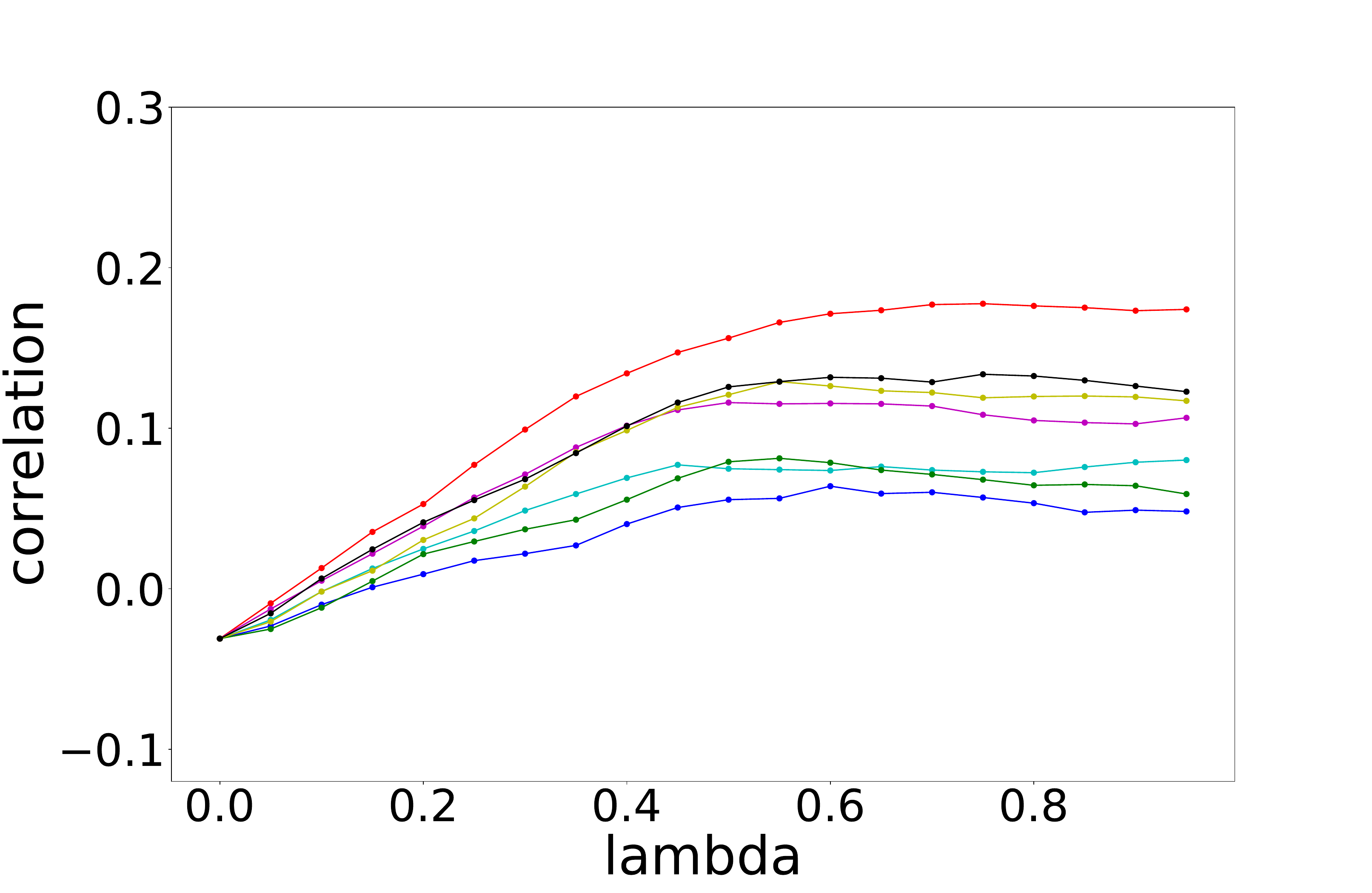}
              \caption{LOR - Consistency}
        \end{subfigure}
        \hfill%
        \begin{subfigure}[b]{0.32\textwidth}
              \includegraphics[width=\linewidth,trim={10 10 155 105},clip]{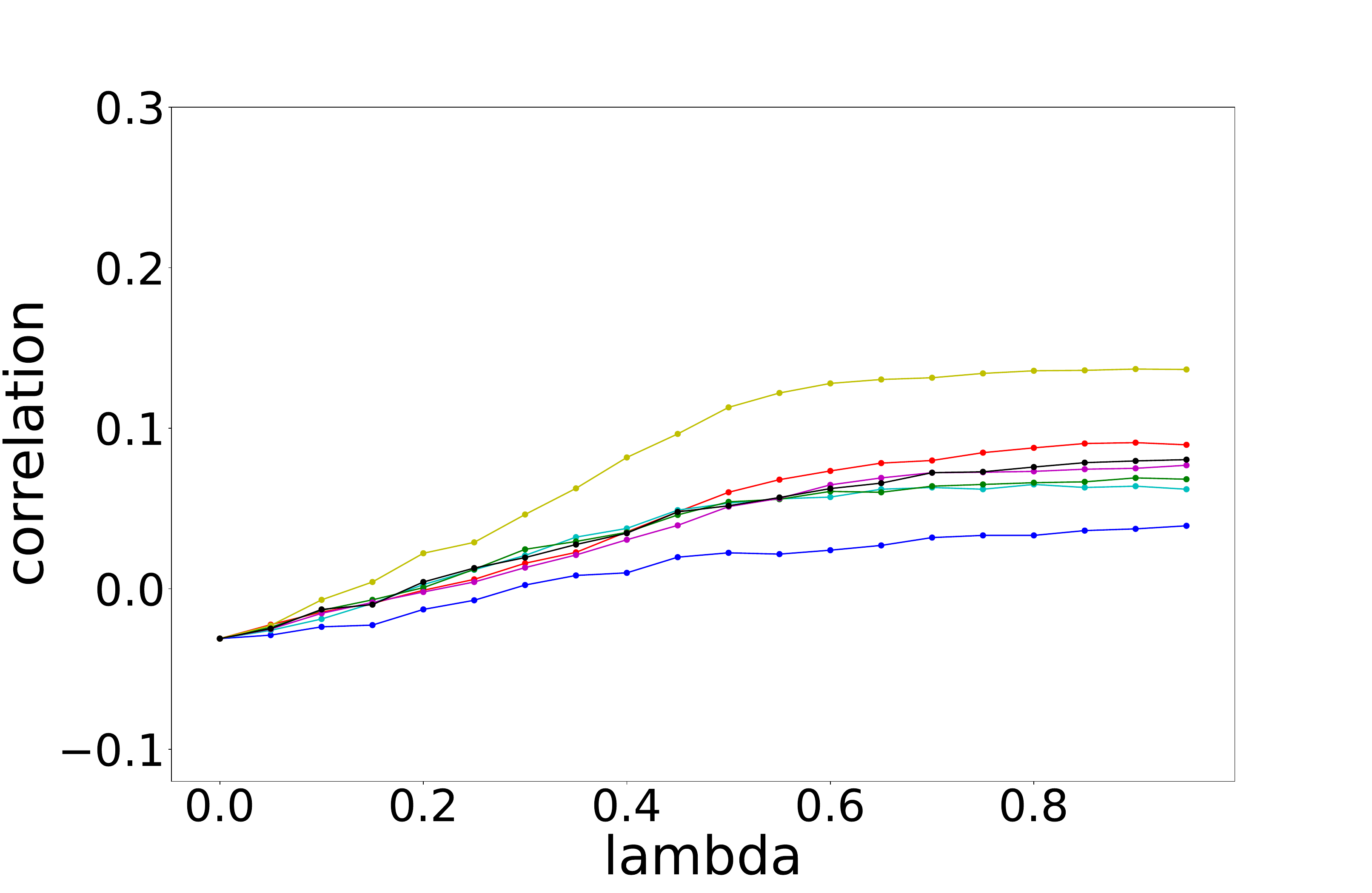}
            \caption{MOR - Consistency}
        \end{subfigure}
        \hfill%
        \begin{subfigure}[b]{0.32\textwidth}
              \includegraphics[width=\linewidth,trim={10 10 155 105},clip]{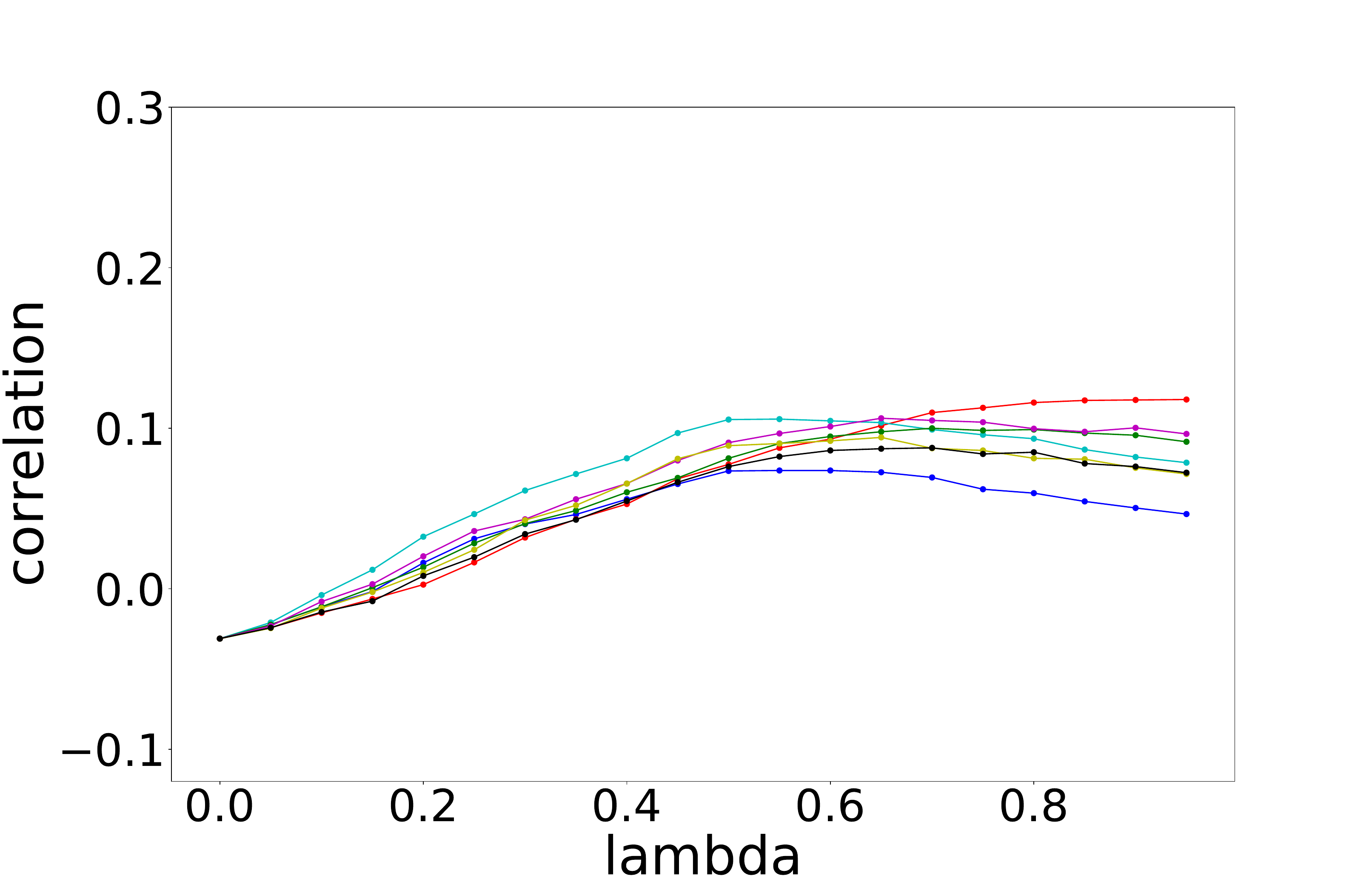}
              \caption{HOR - Consistency}
        \end{subfigure}
        
        \begin{subfigure}[b]{0.32\textwidth}
              \includegraphics[width=\linewidth,trim={10 10 155 105},clip]{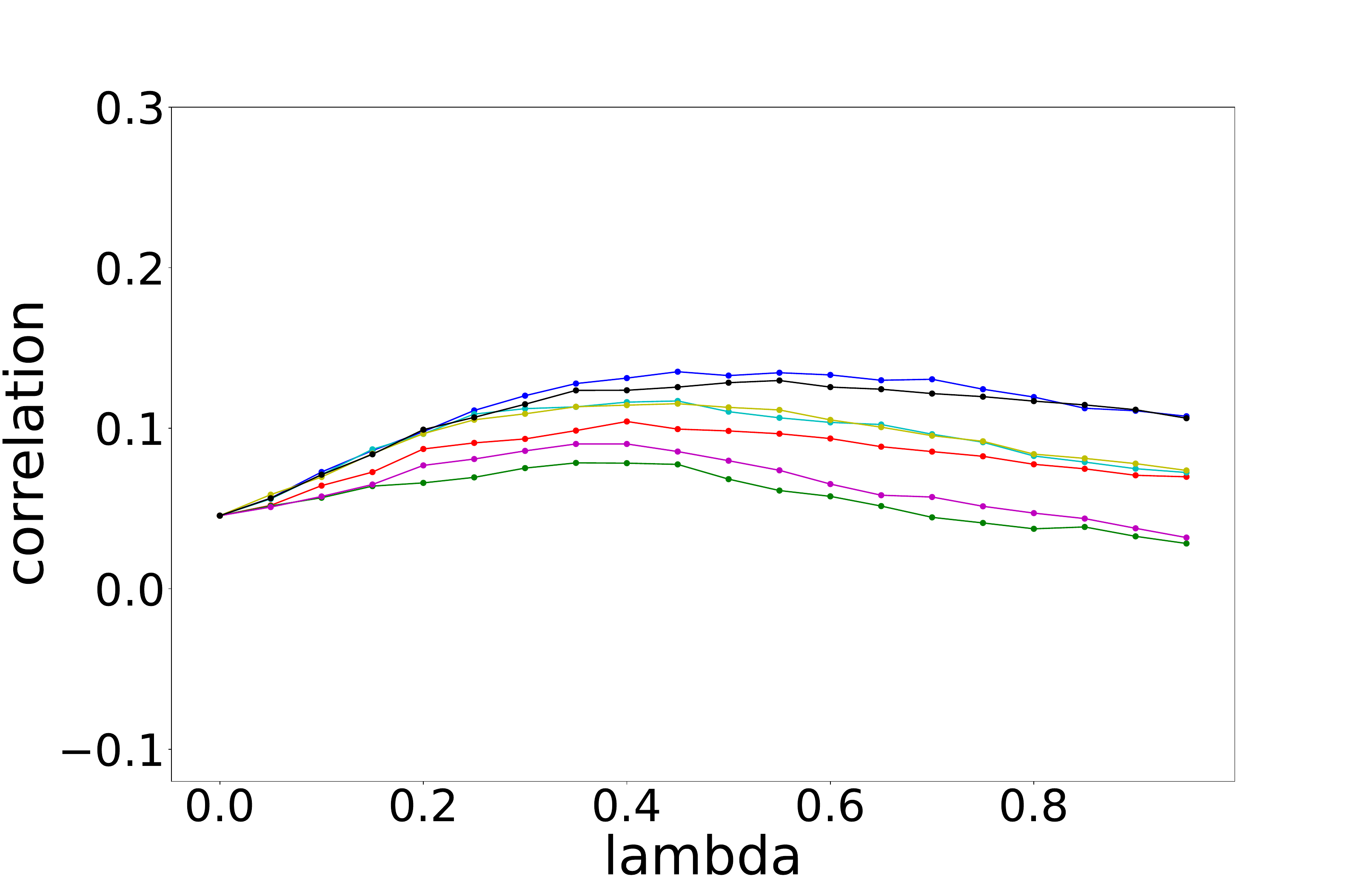}
              \caption{LOR - Relevance}
        \end{subfigure}
        \hfill%
        \begin{subfigure}[b]{0.32\textwidth}
              \includegraphics[width=\linewidth,trim={10 10 155 105},clip]{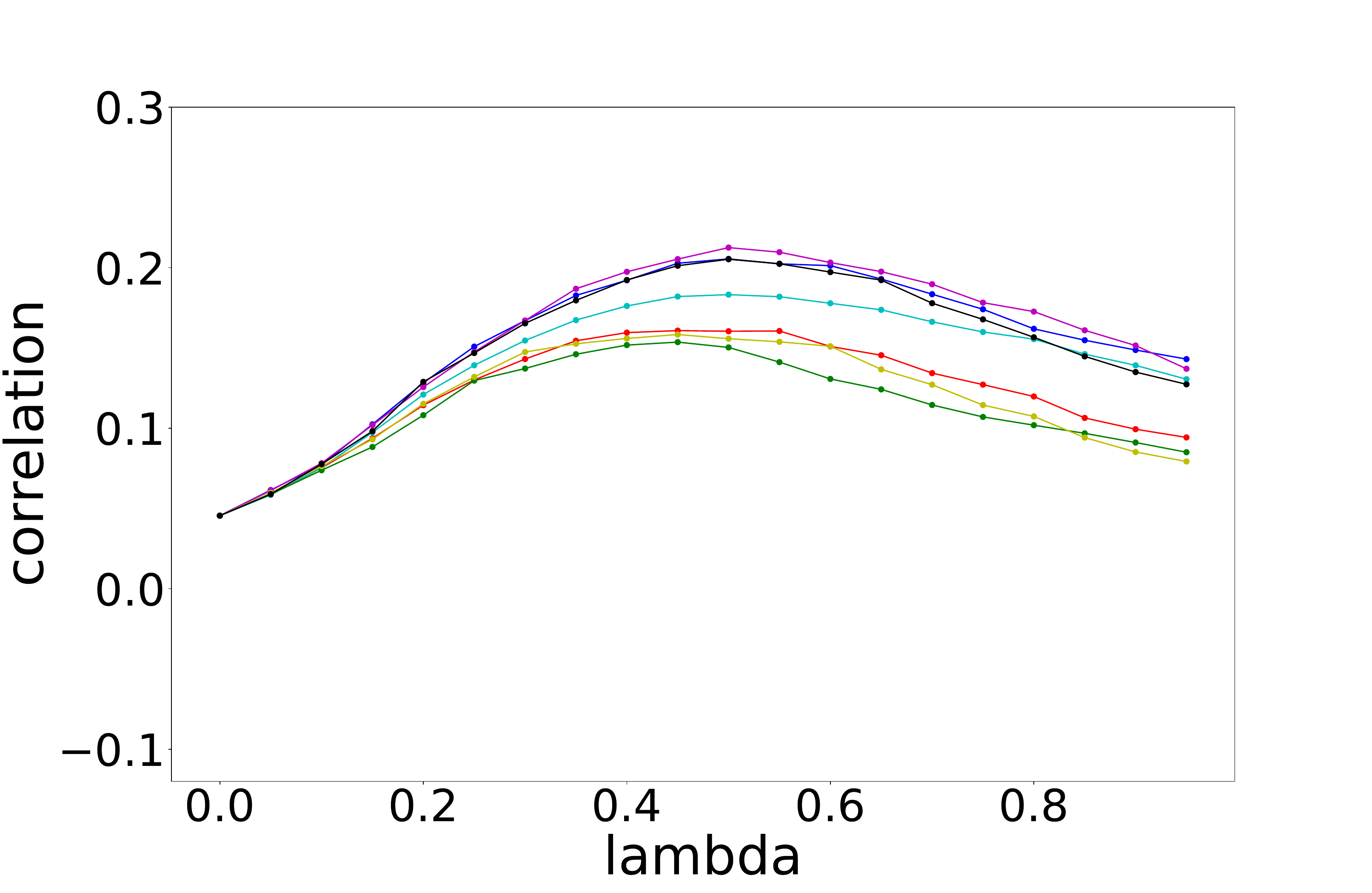}
            \caption{MOR - Relevance}
        \end{subfigure}
        \hfill%
        \begin{subfigure}[b]{0.32\textwidth}
              \includegraphics[width=\linewidth,trim={10 10 155 105},clip]{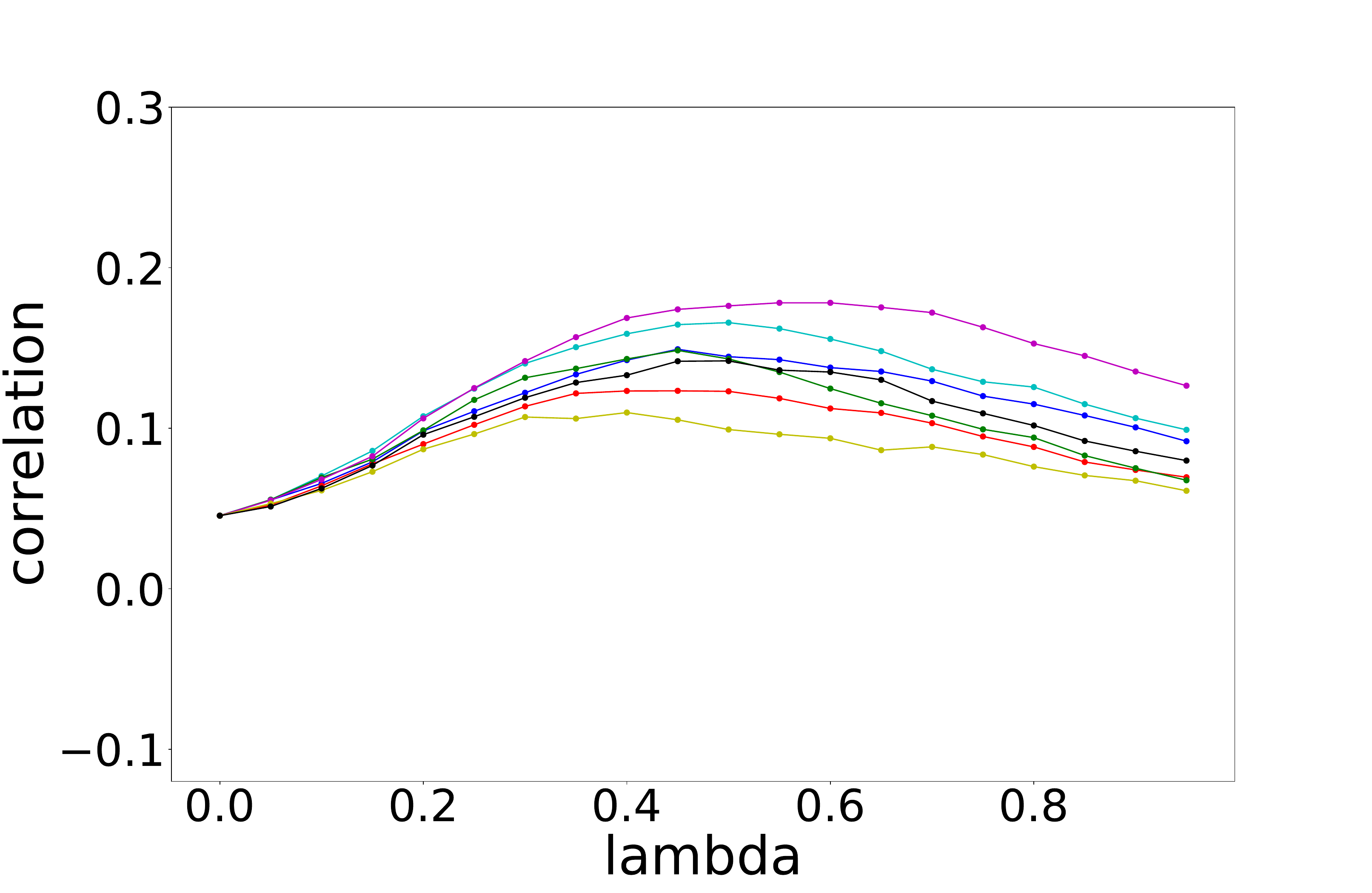}
              \caption{HOR - Relevance}
        \end{subfigure}
        
        \begin{subfigure}[b]{0.32\textwidth}
              \includegraphics[width=\linewidth,trim={10 10 155 105},clip]{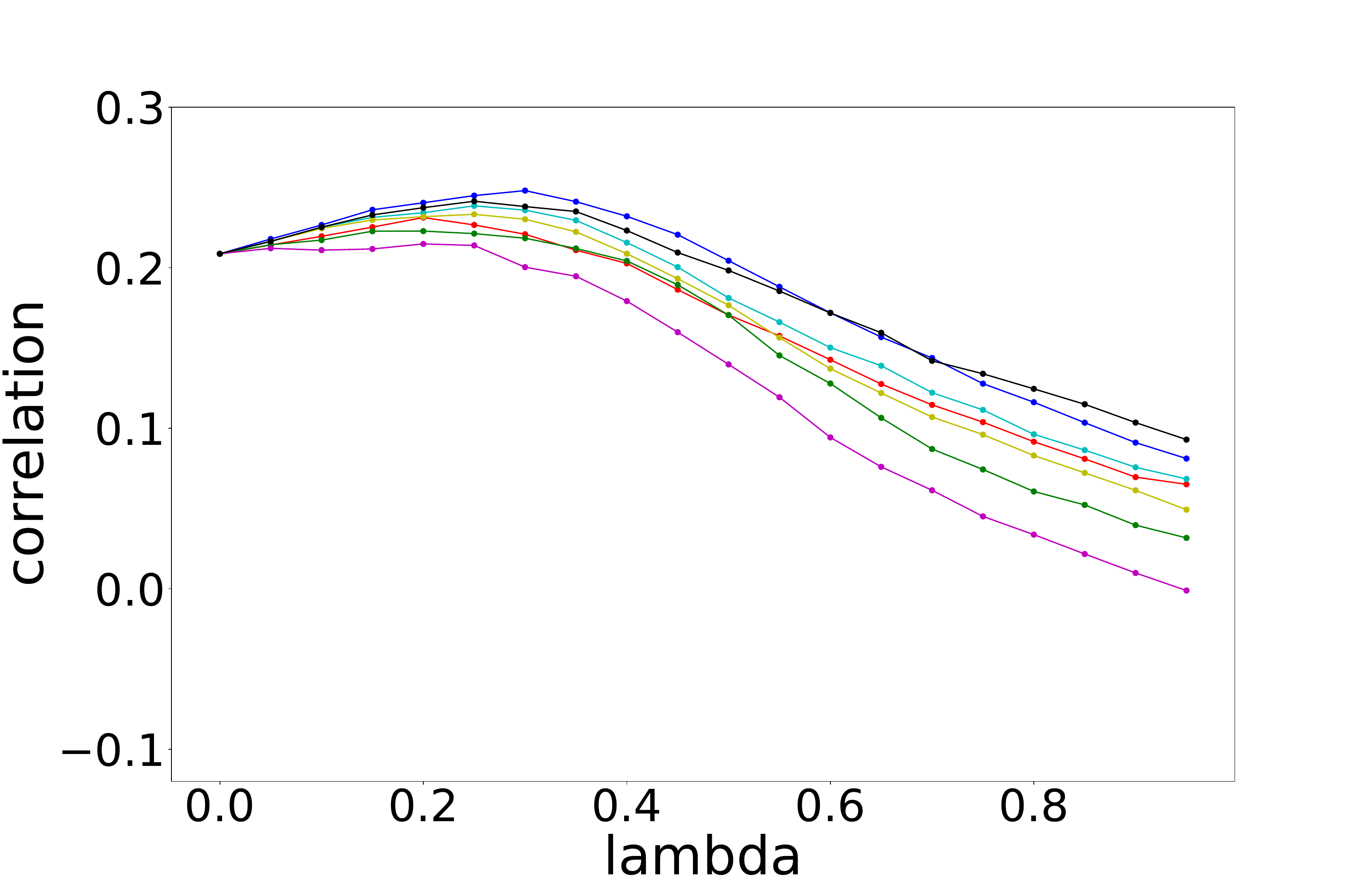}
              \caption{LOR - Coherence}
        \end{subfigure}
        \hfill%
        \begin{subfigure}[b]{0.32\textwidth}
              \includegraphics[width=\linewidth,trim={10 10 155 105},clip]{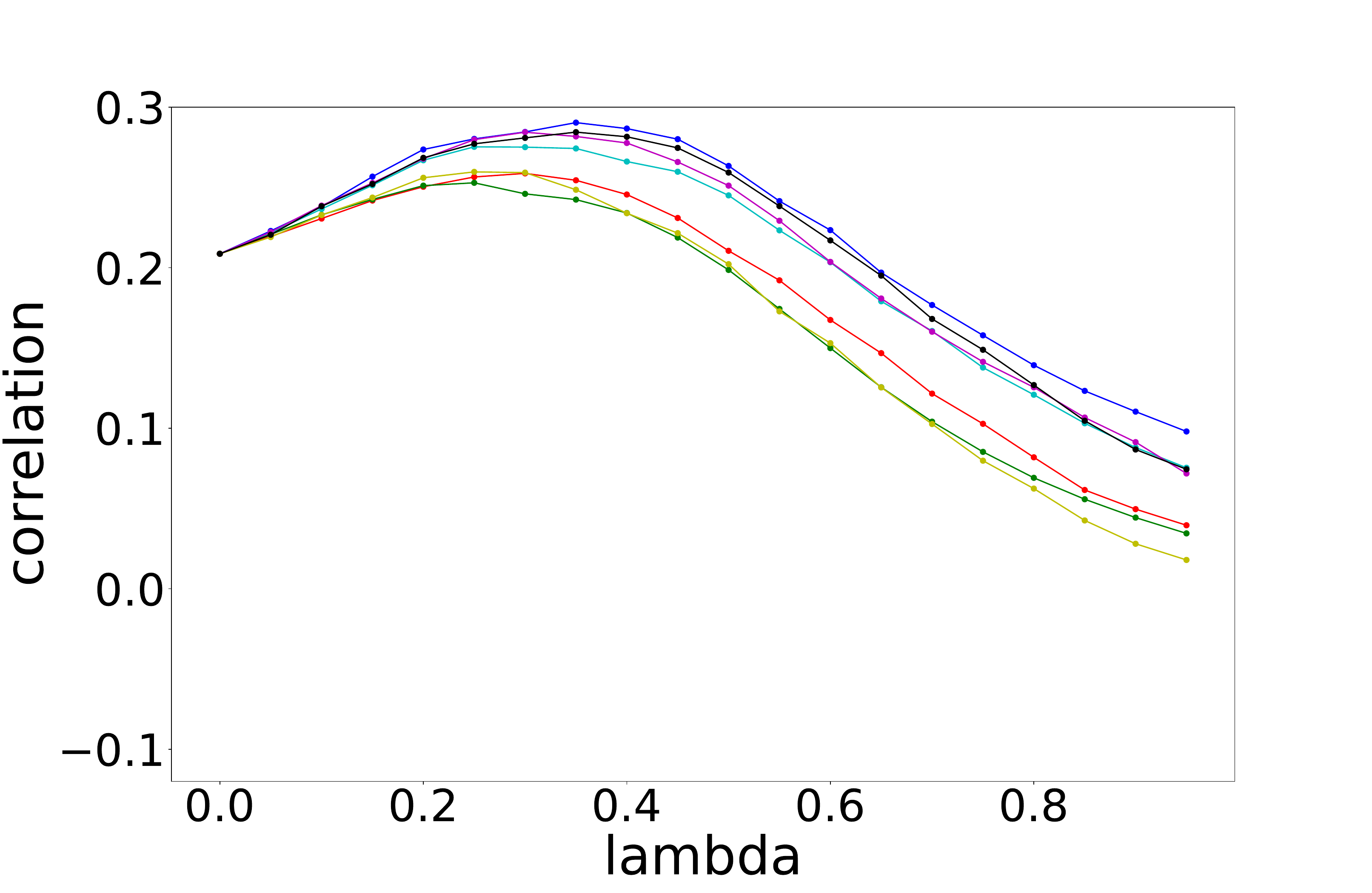}
            \caption{MOR - Coherence}
        \end{subfigure}
        \hfill%
        \begin{subfigure}[b]{0.32\textwidth}
              \includegraphics[width=\linewidth,trim={10 10 155 105},clip]{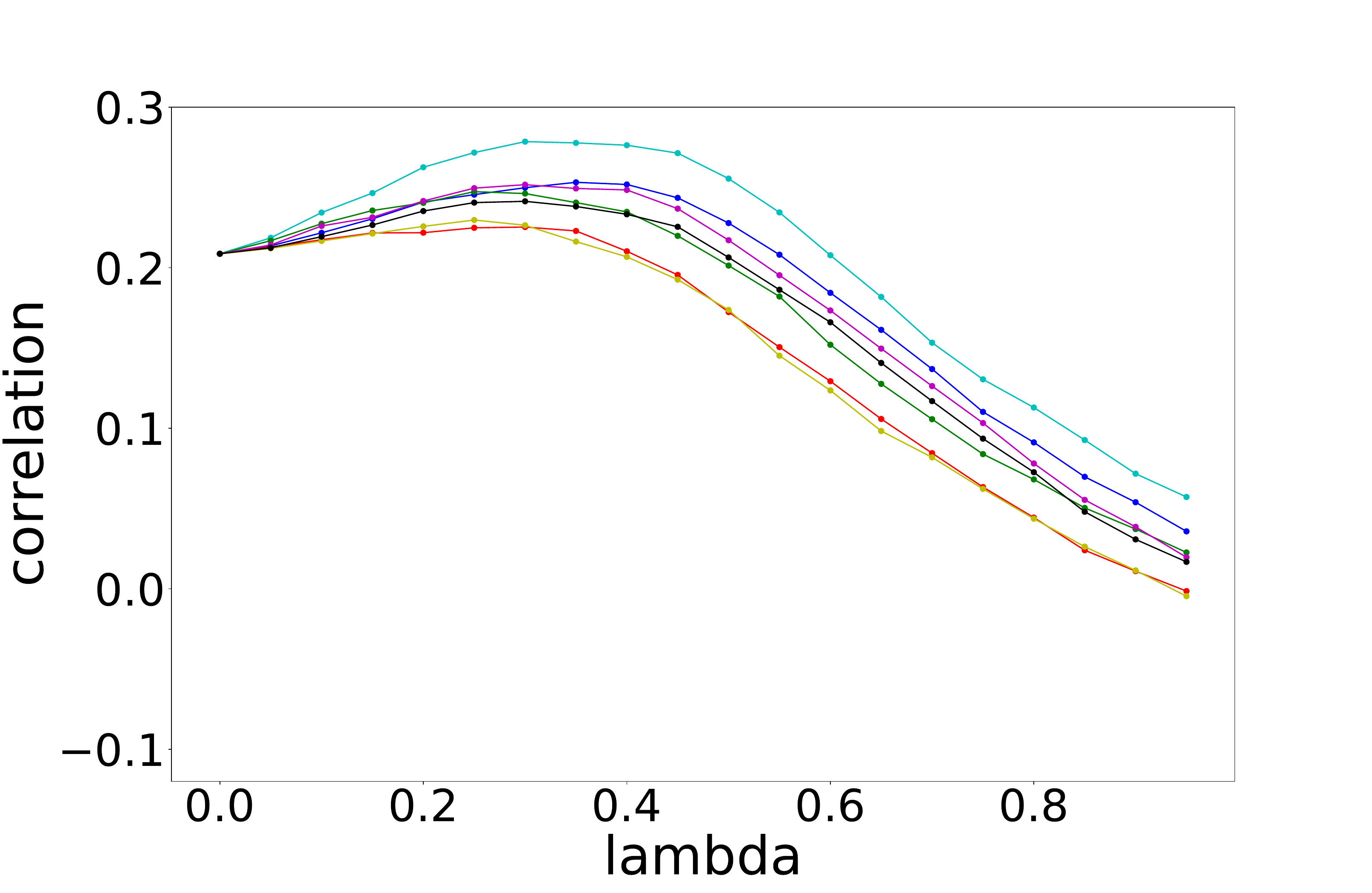}
              \caption{HOR - Coherence}
        \end{subfigure}
        
        \begin{subfigure}[b]{0.32\textwidth}
              \includegraphics[width=\linewidth,trim={10 10 155 105},clip]{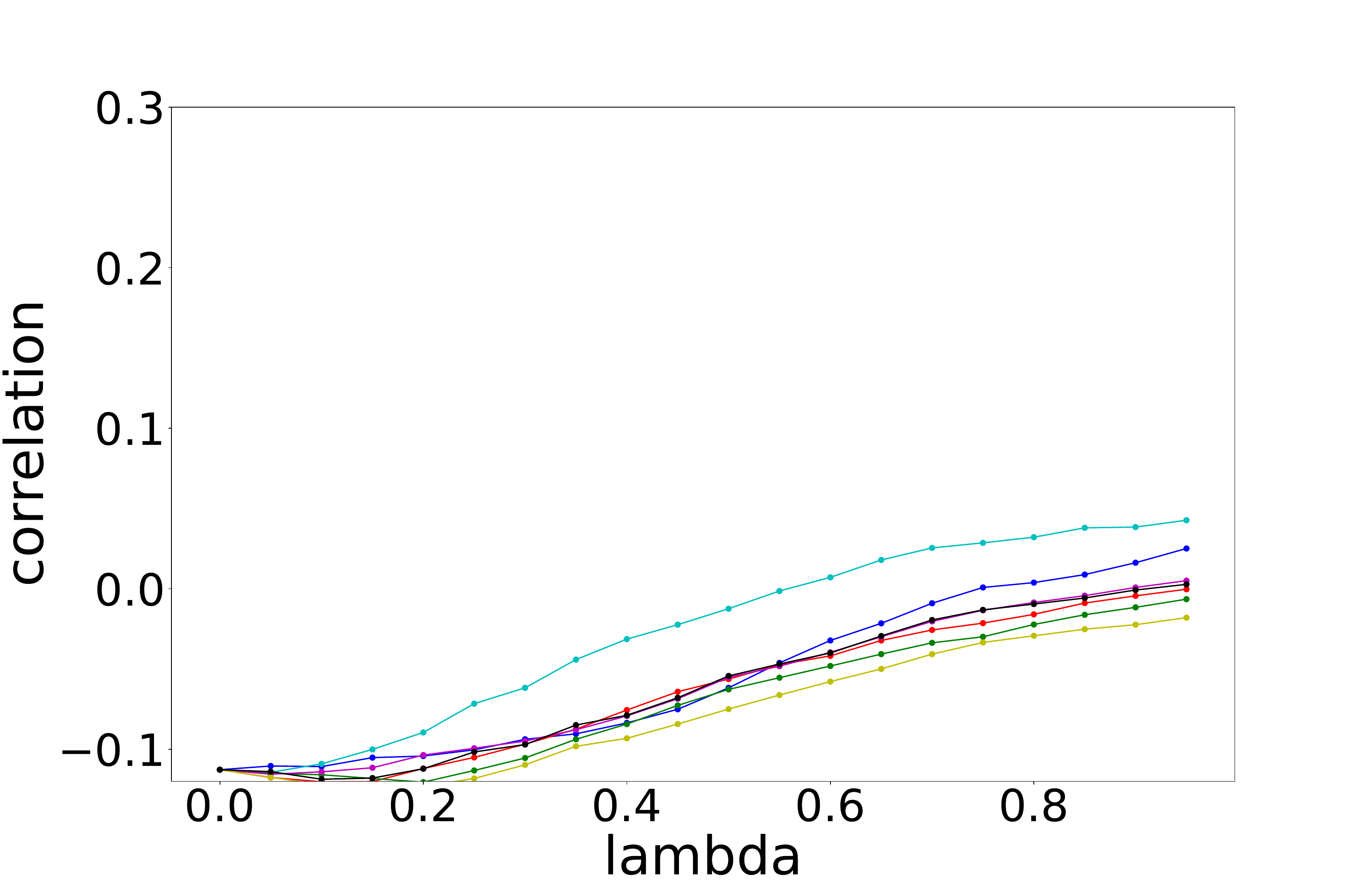}
              \caption{LOR - Fluency}
        \end{subfigure}
        \hfill%
        \begin{subfigure}[b]{0.32\textwidth}
              \includegraphics[width=\linewidth,trim={10 10 155 105},clip]{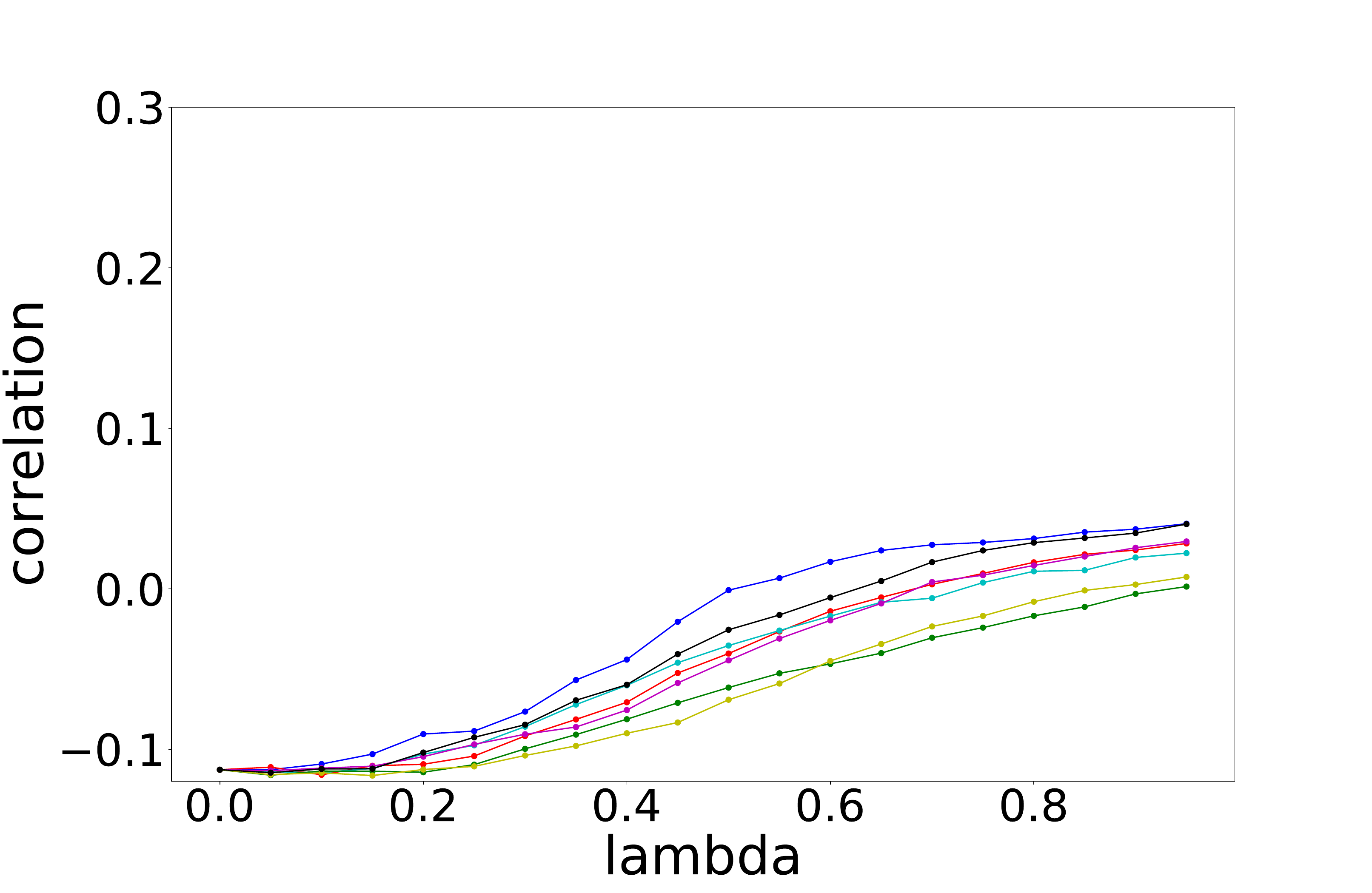}
            \caption{MOR - Fluency}
        \end{subfigure}
        \hfill%
        \begin{subfigure}[b]{0.32\textwidth}
              \includegraphics[width=\linewidth,trim={10 10 155 105},clip]{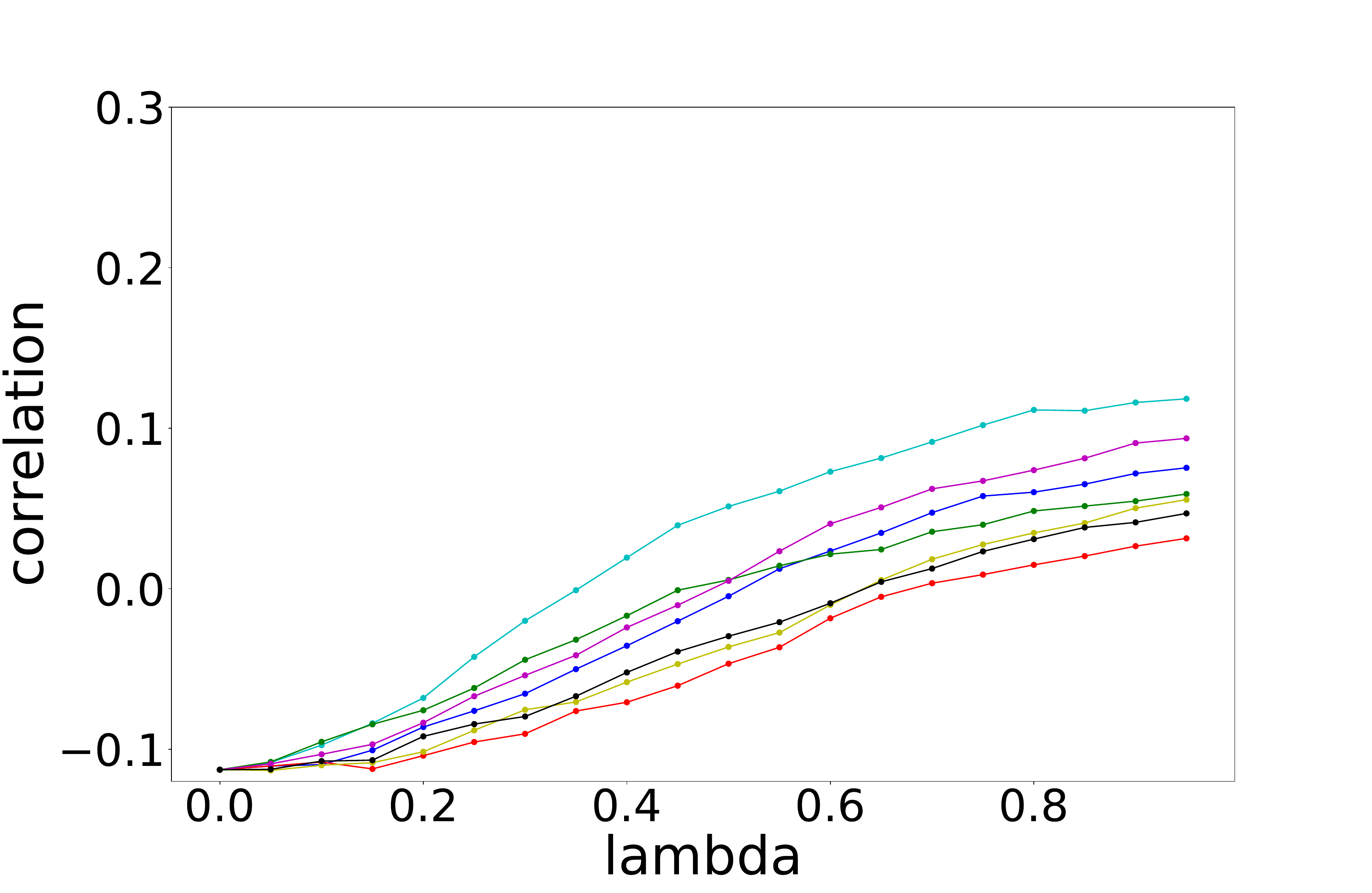}
              \caption{HOR - Fluency}
        \end{subfigure}

        \caption{Kendall Tau ($\tau$) Correlation coefficient when lambda ($\lambda)$ $\in [0, 1]$ from (a)-(c) for Consistency, (d)-(f) for relevance, (g)-(i) for coherence and (j)-(l) for Fluency dimension when ROUGE score is used as redundancy penalty for less overlapping reference (LOR), medium overlapping reference (MOR) and high overlapping reference (HOR).}
        \label{fig:rouge}
\end{figure*}

\begin{table*}[!htb]\footnotesize
\centering
\resizebox{\textwidth}{!}{
\begin{tabular}{ll}
\hline
\multicolumn{2}{l}{\begin{minipage}[t]{\textwidth}%
\textbf{Article:} Last week she was barely showing – but Demelza Poldark is now the proud mother to the show’s latest addition. Within ten minutes of tomorrow night’s episode, fans will see Aidan Turner’s dashing Ross Poldark gaze lovingly at his new baby daughter. As Sunday night’s latest heartthrob, women across the country have voiced their longing to settle down with the brooding Cornish gentleman – but unfortunately, it seems as if his heart is well and truly off the market. Scroll down for the video. Last week she was barely showing – but Demelza Poldark is now the proud mother to the show’s latest addition He may have married his red-headed kitchen maid out of duty, but as he tells her that she makes him a better man, audiences can have little doubt about his feelings. What is rather less convincing, however, is the timeline of the pregnancy. With the climax of the previous episode being the announcement of the pregnancy, it is quite a jump to the start of tomorrow’s installment where Demelza, played by Eleanor Tomlinson, talks about being eight months pregnant. Just minutes after – once again without any nod to the passing of time – she is giving birth, with the last month of her pregnancy passing in less than the blink of an eye. With the climax of the previous episode being the announcement of the pregnancy, it is quite a jump to the start of tomorrow’s instalment where Demelza, played by Eleanor Tomlinson, talks about being eight months pregnant As Sunday night’s latest heartthrob, women across the country have voiced their longing to settle down with Poldark – but unfortunately, it seems as if his heart is well and truly off the market Their fast relationship didn't go unnoticed by fans. One posted on Twitter: ‘If you are pregnant in Poldark times expect to have it in the next 10 minutes’ It is reminiscent of the show’s previous pregnancy that saw Elizabeth, another contender for Ross’s affection, go to full term in the gap between two episodes. This didn’t go unnoticed by fans, who posted on Twitter: ‘Poldark is rather good, would watch the next one now. Though if you are pregnant in Poldark times expect to have it in the next 10 minutes. \vspace{1 mm}%
\end{minipage}} \\ \hline

% \multicolumn{1}{c}{\textbf{LOR}} & \multicolumn{1}{c}{\textbf{MOR}} \\ \hline
\multicolumn{1}{l}{\begin{minipage}[t]{\textwidth}%
\textbf{Model Summary:} Within ten minutes of tomorrow night's episode, fans will see aidan turner's dashing ross poldark gaze lovingly at his new baby daughter. Last week she was barely showing -- but demelza poldark is now the proud mother to the show's latest addition. Last week she was barely showing -- but demelza poldark is now the proud mother to the show's latest addition. \textcolor{red}{(clearly redundant extractive summary)}%
\end{minipage}}  \\ \hline

\multicolumn{2}{l}{\begin{minipage}[t]{\textwidth}%
\textbf{Score\textsubscript{red} for model summary}: 0.40 \vspace{1 mm}%
\end{minipage}} \\ \hline

\multicolumn{1}{l}{\begin{minipage}[t]{\textwidth}%
\textbf{Less Overlapping Reference (LOR)}: A celebrity recently welcomed a baby into the world and the wife discusses her experiences with her pregnancy. She has wanted to settle down for a while and is glad her pregnancy wasn't noticeable on television.%
\end{minipage}}\\ \hline

\multicolumn{1}{l}{\begin{minipage}[t]{\textwidth}%
\textbf{Medium Overlapping/CNN Reference (MOR)}: SPOILER ALERT: Maid gives birth to baby on Sunday's episode. Only announced she was pregnant with Poldark's baby last week.
\end{minipage}}\\ \hline

\multicolumn{1}{l}{\begin{minipage}[t]{\textwidth}%
\textbf{High Overlapping Reference (HOR)}: In the latest episode, Demelza Poldark talks about being 8 months pregnant. Ross Poldark, who is off the market and in love with Demelza, will be shown gazing lovingly at his new baby daughter tomorrow night.
\end{minipage}} \\ \hline

\multicolumn{2}{l}{\begin{minipage}[t]{\textwidth}%
\textbf{Sem-nCG Score} only according to equation~\ref{equ:nCG} for\\
LOR: 0.67 \hspace{5 mm} MOR: 0.733 \hspace{5 mm} HOR: 0.8 \vspace{1 mm}%
\end{minipage}} \\ \hline

\multicolumn{2}{l}{\begin{minipage}[t]{\textwidth}%
\textbf{Revised Sem-nCG Score} along with Score\textsubscript{red} according to equation~\ref{equ:final} for\\
LOR: 0.532 \hspace{5 mm} MOR: 0.565 \hspace{5 mm} HOR: 0.599 \vspace{1 mm}%
\end{minipage}} \\ \hline

\multicolumn{2}{l}{\begin{minipage}[t]{\textwidth}%
\textbf{Human Evaluation} (annotated by experts and score ranged between 0-1) \\
Coherence: 0.47 \hspace{5 mm} Consistency: 1 \hspace{5 mm} Fluency: 1 \hspace{5 mm} Relevance: 0.67%
\end{minipage}} \\ \hline
\end{tabular}
}
\caption{An example of the model summary evaluation using the redundancy-aware Sem-nCG metric.}
\label{tab:ex_sce}
\end{table*}

\section{Results}\label{result}
\subsection{Redundancy-aware \textit{Sem-nCG}}
\label{sub:red_improve}
We first considered how redundancy-aware Sem-nCG performs in extractive summarization with single reference. As shown in Table~\ref{tab:hum}, we computed Kendall's tau ($\tau$) correlation between the expert given score for model summary and the Sem-nCG score with/without redundancy along the four meta-evaluation criteria: \textit{Consistency}, \textit{Relevance}, \textit{Coherence}, and \textit{Fluency}, for different embedding variations (to create the groudtruth ranking) and different approaches to compute \textit{Score\textsubscript{red}}. We utilized Equation~\ref{equ:final} to compute the redundancy-aware \textit{Sem-nCG} score, where lambda ($\lambda$) is a hyper-parameter choice and is set to $\lambda = 0.5$ empirically. In Table~\ref{tab:hum} w/o redundancy refers to Equation~\ref{equ:nCG}.

Table~\ref{tab:hum} shows that the redundancy-aware \textit{Sem-nCG} metric outperforms the original \textit{Sem-nCG} metric in terms of \textit{Consistency}, \textit{Relevance}, and \textit{Coherence}; with a $5\%$ improvement in \textit{Relevance} and a $14\%$ improvement in \textit{Coherence} for less overlapping references (LOR). We also observe improvements in the \textit{Relevance} ($9\%$) and \textit{Coherence} ($20\%$) dimensions for medium overlapping references (MOR). For High Overlapping References (HOR), the improvement is $8\%$ and $22\%$ for \textit{Relevance} and \textit{Coherence}, respectively.

We also observe that STSb-distilbert embedding is a better choice in the \textit{Consistency} dimension, whereas USE with enc-2 is a better choice in the \textit{Relevance} and \textit{Coherence} dimensions to construct the groundtruth ranking. Therefore, we recommend STSb-distilbert to create groundtruth ranking if \textit{Consistency} is a top priority, otherwise, we recommend using USE with enc-2. A groundtruth ranking was also created by combining STSb-distilbert and USE into an ensemble, which showed balanced performance across all four dimensions. It also appears that ROUGE and BERTScore provide comparable performances while computing \textit{Score\textsubscript{red}}. However, using ROUGE score as self-referenced redundancy will be a better choice as evident from Section~\ref{sec:multi_ref}.
 
In Table~\ref{tab:hum_rouge_bert} Kendall's tau correlation of ROUGE and BERTScore has been demonstrated to get an idea of the advantage of redundancy-aware \textit{Sem-nCG} and it is clearly evident that redundancy-aware Sem-nCG also exhibits stronger correlation than these metrics.

\subsection{Hyperparameter Choice} \label{sec:hyper}
In figure~\ref{fig:rouge}, we have varied $\lambda \in [0, 1]$ for the 3 scenarios (LOR, MOR and HOR) and computed human correlation along four dimensions (\textit{Consistency}, \textit{Relevance}, \textit{Coherence} and \textit{Fluency}) when different embeddings are used to create the groundtruth ranking and  ROUGE score is used to compute \textit{Score\textsubscript{red}}. Human correlations with BERTScore-based redundancy are presented in Appendix. For both redundancy penalties, it shows that higher lambda ($\lambda \geq 0.6$) achieves better correlation for the \textit{Consistency} dimensions, which makes sense because higher lambda means giving more weight to \textit{Sem-nCG}. For \textit{Relevance} and \textit{Coherence} dimensions, a lower lambda ($\lambda$) value between $[0.3-0.5]$ is a better choice as lower $\lambda$ means more penalty to redundancy. It appears that for \textit{Fluency} all metric variations struggle. It is evident that $\lambda = 0.5$ gives comparable performance in all four quality dimensions (consistency, relevance, coherence and fluency) and thus we recommend using $\lambda = 0.5$ while adopting Equation~\ref{equ:final} to compute redundancy-aware \textit{Sem-nCG}. Table~\ref{tab:ex_sce} shows a qualitative example for the evaluation of a model-extracted summary.

\begin{table*}[!htb]\large
\centering
\resizebox{\textwidth}{!}{
\begin{tabular}{l|cccc|cccc|cccc|cccc}
\hline
\multicolumn{1}{c}{\multirow{2}{*}{Metric}} & \multicolumn{4}{|c}{\textbf{Multi-Ref LOR, MOR, HOR}} & \multicolumn{4}{|c}{\textbf{Multi-Ref LORs}} & \multicolumn{4}{|c|}{\textbf{Multi-Ref MORs}} & \multicolumn{4}{c}{\textbf{Multi-Ref HORs}} \\ \cline{2-17} 
\multicolumn{1}{c}{} & \multicolumn{1}{|c}{Con} & \multicolumn{1}{c}{Rel} & \multicolumn{1}{c}{Coh} & \multicolumn{1}{c}{Flu} & \multicolumn{1}{|c}{Con} & \multicolumn{1}{c}{Rel} & \multicolumn{1}{c}{Coh} & \multicolumn{1}{c}{Flu} & \multicolumn{1}{|c}{Con} & \multicolumn{1}{c}{Rel} & \multicolumn{1}{c}{Coh} & \multicolumn{1}{c}{Flu} & \multicolumn{1}{|c}{Con} & \multicolumn{1}{c}{Rel} & \multicolumn{1}{c}{Coh} & \multicolumn{1}{c}{Flu} \\ \hline
ROUGE-1 & 0.00 & -0.01 & -0.09 & -0.01 & -0.05 & 0.05 & 0.00 & 0.01 & -0.05 & 0.09 & 0.04 & -0.01 & -0.02 & 0.21 & 0.13 & 0.10 \\
ROUGE-L & 0.00 & -0.01 & -0.09 & -0.01 & 0.00 & 0.04 & -0.01 & 0.01 & -0.06 & 0.07 & 0.04 & 0.00 & -0.01 & 0.15 & 0.09 & -0.04 \\
BERTScore & 0.09 & 0.19 & 0.14 & 0.03 & 0.01 & 0.07 & -0.01 & 0.04 & -0.04 & 0.05 & 0.03 & 0.05 & 0.04 & 0.20 & 0.12 & 0.06 \\ \hline
\end{tabular}}
\caption{Kendall Tau ($\tau$) correlation coefficient for ROUGE and BERTScore for consistency (con), relevance (rel), coherence (coh) and fluency (flu) dimension for evaluating extractive model summaries with multiple references.}
\label{tab:hum_multi_ref_metric}
\end{table*}

\begin{table*}[!htb]\large
\centering
\resizebox{\textwidth}{!}{
\begin{tabular}{l|cccc|cccc|cccc}
\hline
\multicolumn{13}{c}{\textbf{Multi-Ref LOR, MOR, HOR}} \\ \hline
\multicolumn{1}{c|}{\multirow{2}{*}{\textbf{Embedding}}} & \multicolumn{4}{c|}{\textbf{w/o Redundancy}} & \multicolumn{4}{c}{\textbf{+ Redundancy Penalty (ROUGE)}} & \multicolumn{4}{|c}{\textbf{+ Redundancy Penalty (BERTScore)}} \\ \cline{2-13} 
\multicolumn{1}{c|}{} & \multicolumn{1}{l}{Consistency} & \multicolumn{1}{l}{Relevance} & \multicolumn{1}{l}{Coherence} & \multicolumn{1}{l}{Fluency} & \multicolumn{1}{|l}{Consistency} & \multicolumn{1}{l}{Relevance} & \multicolumn{1}{l}{Coherence} & \multicolumn{1}{l}{Fluency} & \multicolumn{1}{|l}{Consistency} & \multicolumn{1}{l}{Relevance} & \multicolumn{1}{l}{Coherence} & \multicolumn{1}{l}{Fluency} \\ \hline
Infersent & 0.07 & 0.11 & 0.08 & \cellcolor{green!15}{\textbf{0.06}} & 0.11 & 0.18 &\cellcolor{green!15}{\textbf{ 0.27}} & 0.01 & 0.09 & 0.18 & 0.20 & 0.03 \\ 
Elmo & 0.09 & 0.06 & 0.01 & 0.00 & 0.09 & 0.12 & 0.18 & -0.05 & 0.09 & 0.12 & 0.11 & -0.03 \\
STSb-bert & 0.10 & 0.12 & 0.04 & 0.06 & 0.09 & 0.19 & 0.24 & -0.02 & 0.10 & \cellcolor{green!15}{\textbf{0.20}} & 0.18 & 0.01 \\
STSb-roberta & 0.14 & 0.10 & 0.01 & 0.02 & 0.12 & 0.17 & 0.21 & -0.06 & \cellcolor{green!15}{\textbf{0.13}} & 0.17 & 0.13 & -0.02 \\
USE & 0.04 & 0.12 & 0.08 & 0.05 & 0.06 & 0.19 & 0.26 & -0.03 & 0.05 & 0.19 & 0.20 & 0.01 \\ 
STSb-distilbert & 0.14 & 0.13 & 0.05 & 0.02 & 0.10 & 0.19 & 0.23 & -0.04 & 0.12 & \cellcolor{green!15}{\textbf{0.20}} & 0.17 & -0.01\\ \hline \hline
\multicolumn{13}{c}{\textbf{Multi-Ref LORs}} \\ \hline
\multicolumn{1}{c|}{\multirow{2}{*}{\textbf{Embedding}}} & \multicolumn{4}{c|}{\textbf{w/o Redundancy}} & \multicolumn{4}{c}{\textbf{+ Redundancy Penalty (ROUGE)}} & \multicolumn{4}{|c}{\textbf{+ Redundancy Penalty (BERTScore)}} \\ \cline{2-13} 
\multicolumn{1}{c|}{} & \multicolumn{1}{l}{Consistency} & \multicolumn{1}{l}{Relevance} & \multicolumn{1}{l}{Coherence} & \multicolumn{1}{l}{Fluency} & \multicolumn{1}{|l}{Consistency} & \multicolumn{1}{l}{Relevance} & \multicolumn{1}{l}{Coherence} & \multicolumn{1}{l}{Fluency} & \multicolumn{1}{|l}{Consistency} & \multicolumn{1}{l}{Relevance} & \multicolumn{1}{l}{Coherence} & \multicolumn{1}{l}{Fluency} \\ \hline
Infersent & 0.03 & 0.10 & 0.09 & \cellcolor{green!15}{\textbf{0.07}} & 0.05 & \cellcolor{green!15}{\textbf{0.16}} & \cellcolor{green!15}{\textbf{0.25}} & 0.02 & 0.02 & 0.15 & 0.18 & 0.04 \\
Elmo & 0.04 & 0.05 & -0.04 & 0.03 & 0.05 & 0.12 & 0.15 & -0.04 & 0.03 & 0.11 & 0.06 & -0.01 \\
STSb-bert & 0.08 & 0.10 & 0.02 & 0.01 & 0.09 & 0.15 & 0.20 & -0.06 & 0.06 & 0.15 & 0.13 & -0.04 \\
STSb-roberta & 0.10 & 0.07 & -0.04 & 0.00 & \cellcolor{green!15}{\textbf{0.11}} & 0.15 & 0.17 & -0.07 & 0.09 & 0.15 & 0.09 & -0.04 \\
USE & 0.02 & 0.05 & 0.01 & 0.03 & 0.04 & 0.12 & 0.19 & -0.04 & 0.02 & 0.10 & 0.12 & 0.00 \\
STSb-distilbert & 0.10 & 0.04 & -0.02 & -0.02 & \cellcolor{green!15}{\textbf{0.11}} & 0.09 & 0.15 & -0.09 & 0.09 & 0.09 & 0.09 & -0.07 \\ \hline
\hline
\multicolumn{13}{c}{\textbf{Multi-Ref MORs}} \\ \hline
\multicolumn{1}{c|}{\multirow{2}{*}{\textbf{Embedding}}} & \multicolumn{4}{c|}{\textbf{w/o Redundancy}} & \multicolumn{4}{c}{\textbf{+ Redundancy Penalty (ROUGE)}} & \multicolumn{4}{|c}{\textbf{+ Redundancy Penalty (BERTScore)}} \\ \cline{2-13} 
\multicolumn{1}{c|}{} & \multicolumn{1}{l}{Consistency} & \multicolumn{1}{l}{Relevance} & \multicolumn{1}{l}{Coherence} & \multicolumn{1}{l}{Fluency} & \multicolumn{1}{|l}{Consistency} & \multicolumn{1}{l}{Relevance} & \multicolumn{1}{l}{Coherence} & \multicolumn{1}{l}{Fluency} & \multicolumn{1}{|l}{Consistency} & \multicolumn{1}{l}{Relevance} & \multicolumn{1}{l}{Coherence} & \multicolumn{1}{l}{Fluency} \\ \hline
Infersent & 0.08 & 0.08 & 0.03 & \cellcolor{green!15}{\textbf{0.06}} & 0.10 & \cellcolor{green!15}{\textbf{0.15}} & \cellcolor{green!15}{\textbf{0.23}} & -0.02 & 0.08 & \cellcolor{green!15}{\textbf{0.15}} & 0.16 & 0.02 \\
Elmo & 0.06 & 0.05 & -0.02 & 0.00 & 0.04 & 0.13 & 0.16 & -0.07 & 0.05 & 0.11 & 0.08 & -0.05 \\
STSb-bert & 0.07 & 0.05 & 0.02 & 0.01 & 0.09 & 0.13 & 0.22 & -0.08 & 0.07 & 0.12 & 0.15 & -0.04 \\
STSb-roberta & 0.05 & 0.07 & -0.01 & 0.02 & 0.07 & 0.14 & 0.21 & -0.07 & 0.04 & 0.14 & 0.14 & -0.03 \\
USE & 0.02 & 0.08 & 0.05 & 0.01 & 0.04 & 0.15 & \cellcolor{green!15}{\textbf{0.25}} & -0.06 & 0.02 & 0.14 & 0.17 & -0.03 \\
STSb-distilbert & \cellcolor{green!15}{\textbf{0.11}} & 0.01 & 0.00 & -0.01 & 0.09 & 0.07 & 0.17 & -0.10 & 0.10 & 0.06 & 0.10 & -0.05 \\ \hline\hline
\multicolumn{13}{c}{\textbf{Multi-Ref HORs}} \\ \hline
\multicolumn{1}{c|}{\multirow{2}{*}{\textbf{Embedding}}} & \multicolumn{4}{c|}{\textbf{w/o Redundancy}} & \multicolumn{4}{c}{\textbf{+ Redundancy Penalty (ROUGE)}} & \multicolumn{4}{|c}{\textbf{+ Redundancy Penalty (BERTScore)}} \\ \cline{2-13} 
\multicolumn{1}{c|}{} & \multicolumn{1}{l}{Consistency} & \multicolumn{1}{l}{Relevance} & \multicolumn{1}{l}{Coherence} & \multicolumn{1}{l}{Fluency} & \multicolumn{1}{|l}{Consistency} & \multicolumn{1}{l}{Relevance} & \multicolumn{1}{l}{Coherence} & \multicolumn{1}{l}{Fluency} & \multicolumn{1}{|l}{Consistency} & \multicolumn{1}{l}{Relevance} & \multicolumn{1}{l}{Coherence} & \multicolumn{1}{l}{Fluency} \\ \hline
Infersent & 0.07 & 0.08 & 0.05 & 0.03 & \cellcolor{green!15}{\textbf{0.11}} & 0.16 & 0.23 & -0.02 & 0.07 & 0.15 & 0.15 & 0.01 \\
Elmo & 0.04 & 0.09 & 0.02 & 0.06 & 0.06 & 0.16 & 0.19 & 0.00 & 0.04 & 0.14 & 0.11 & 0.03 \\
STSb-bert & 0.08 & 0.11 & 0.04 & 0.05 & 0.12 & 0.18 & \cellcolor{green!15}{\textbf{0.24}} & -0.03 & 0.08 & 0.18 & 0.16 & 0.01 \\
STSb-roberta & 0.10 & 0.09 & 0.01 & 0.03 & 0.14 & 0.17 & 0.22 & -0.04 & 0.10 & 0.16 & 0.13 & 0.00 \\
USE & 0.04 & 0.14 & 0.07 & 0.05 & 0.07 & 0.20 & \cellcolor{green!15}{\textbf{0.24}} & -0.03 & 0.04 & \cellcolor{green!15}{\textbf{0.21}} & 0.18 & 0.01 \\
STSb-distilbert & 0.08 & 0.09 & 0.02 & 0.05 & \cellcolor{green!15}{\textbf{0.11}} & 0.15 & 0.22 & -0.03 & 0.09 & 0.15 & 0.14 & 0.02\\ \hline

 \hline
\end{tabular}}
\caption{Kendall Tau ($\tau$) correlation coefficient for \textbf{Ensemble\textsubscript{sim}} when lambda ($\lambda)$ $= 0.5$ for consistency, relevance, coherence and fluency dimension without redundancy and when ROUGE and BERTScore is used as redundancy penalty for different terminology variations of multiple references (highly abstractive (LORs), medium overlapping (MORs) and highly extractive (HORs) references). The best value in each dimension has been bold green.} 
\label{tab:hum_multi_ref_en_sim}
\end{table*}

\subsection{Redundancy-aware \textit{Sem-nCG} for Evaluation with Multiple References} \label{sec:multi_ref}
SummEval~\cite{DBLP:journals/tacl/FabbriKMXSR21} dataset contains 11 reference summaries. For summary evaluation with multiple references, we considered the lexical overlap of the reference summaries with the original document to demonstrate the terminology variations. Then we considered 3 less overlapping references as Multi-Ref LORs, 3 medium overlapping references as Multi-Ref MORs and 3 high overlapping references as Multi-Ref HORs. We have also mixed up 1 LOR, 1 MOR and 1 HOR and considered this set as Muti-Ref LOR, MOR, HOR to see how the evaluation metric correlates in different terminology variations. Table~\ref{tab:hum_multi_ref_metric} confirms that ROUGE shows very poor correlation in all the dimensions (consistency, relevance, coherence, and fluency) in all the scenarios and shows slightly better correlation in Multi-Ref HORs (which is somewhat expected as ROUGE considers direct lexical overlap). Interestingly, BERTScore also shows poor correlation in all the settings supporting that the traditional evaluation metric becomes less stable for multiple reference summaries with lots of terminology variations~\cite{DBLP:conf/lrec/CohanG16}.

In the original \textit{Sem-nCG} metric, a groundtruth ranking is prepared by considering the cosine similarity between each sentence of the document and reference summary but the evaluation with multiple-reference was left as future work. As a starting point, how to incorporate multiple-reference summaries in the original \textit{Sem-nCG} metric, we designed how to create the groundtruth ranking by considering multiple references. Here, we took the naive approach, first computing cosine similarity of each sentence of the document with each reference among multiple references. Then average it, which we called Ensemble\textsubscript{sim}. For Ensemble\textsubscript{rel}, for each groundtruth ranking prepared for each reference among multiple reference summaries, we took the average of relevance (as it was computed in previously proposed \textit{Sem-nCG} metric~\cite{DBLP:conf/acl/AkterBS22}) and based on that we merged the groundtruth rankings into one groundtruth ranking. Then we use this groundtruth ranking to compute \textit{Sem-nCG} for model extracted summary. With the original Sem-nCG metric, we have also incorporated redundancy into the \textit{Sem-nCG} metric utilizing equation~\ref{equ:final}. We have only considered ROUGE and BERTScore as redundancy penalty both in Table~\ref{tab:hum_multi_ref_en_sim} and~\ref{tab:hum_multi_ref_en_rel} when $\lambda = 0.5$ (as evident from Section~\ref{sec:hyper} that this setting gives better performance). We have also considered different embedding variations to create the groundtruth ranking.

From Table~\ref{tab:hum_multi_ref_en_sim}, we can see that redundancy-aware \textit{Sem-nCG} shows better correlations for all the scenarios (multi-ref LORs, multi-ref MORs, multi-ref HORs and mixture of LOR, MOR \& HOR). Both ROUGE and BERTScore provide comparable results for self-referenced redundancy penalties, with ROUGE score-based redundancy providing a marginally superior result. Interestingly, redundancy-aware \textit{Sem-nCG} shows robust performance in all the scenarios while showing 25\% improvement in coherence and 10\% improvement in relevance dimension. Same patterns are observed when Ensemble\textsubscript{rel} is also used for the evaluation of multiple reference (See Table~\ref{tab:hum_multi_ref_en_rel}).  

From our empirical evaluation, we would recommend USE embedding to create Ensemble\textsubscript{sim} (merging sentence-wise similarities across different references) with ROUGE redundancy penalty to evaluate extractive summary with multiple references.

\begin{table*}[!htb]\large
\centering
\resizebox{\textwidth}{!}{
\begin{tabular}{l|cccc|cccc|cccc}
\hline
\multicolumn{13}{c}{\textbf{Multi-Ref LOR, MOR, HOR}} \\ \hline
\multicolumn{1}{c|}{\multirow{2}{*}{\textbf{Embedding}}} & \multicolumn{4}{c|}{\textbf{w/o Redundancy}} & \multicolumn{4}{c}{\textbf{+ Redundancy Penalty (ROUGE)}} & \multicolumn{4}{|c}{\textbf{+ Redundancy Penalty (BERTScore)}} \\ \cline{2-13} 
\multicolumn{1}{c|}{} & \multicolumn{1}{l}{Consistency} & \multicolumn{1}{l}{Relevance} & \multicolumn{1}{l}{Coherence} & \multicolumn{1}{l}{Fluency} & \multicolumn{1}{|l}{Consistency} & \multicolumn{1}{l}{Relevance} & \multicolumn{1}{l}{Coherence} & \multicolumn{1}{l}{Fluency} & \multicolumn{1}{|l}{Consistency} & \multicolumn{1}{l}{Relevance} & \multicolumn{1}{l}{Coherence} & \multicolumn{1}{l}{Fluency}\\ \hline
Infersent & 0.09 & 0.10 & 0.04 & 0.08 & 0.11 & 0.17 & 0.24 & 0.01 & 0.09 & 0.18 & 0.18 & 0.04 \\
Elmo & 0.09 & 0.06 & 0.02 & 0.01 & 0.09 & 0.13 & 0.20 & -0.05 & 0.09 & 0.12 & 0.13 & -0.03 \\
STSb-bert & 0.12 & 0.15 & 0.04 & 0.05 & 0.12 & 0.22 & 0.24 & -0.03 & 0.12 & 0.24 & 0.18 & 0.01 \\
STSb-roberta & \cellcolor{green!15}{\textbf{0.14}} & 0.08 & 0.01 & 0.01 & 0.13 & 0.15 & 0.21 & -0.05 & 0.13 & 0.15 & 0.12 & -0.02 \\
USE & 0.04 & 0.16 & 0.11 & \cellcolor{green!15}{\textbf{0.08}} & 0.05 & 0.21 & \cellcolor{green!15}{\textbf{0.29}} & 0.00 & 0.04 & 0.22 & 0.24 & 0.05 \\
STSb-distilbert & 0.14 & 0.10 & 0.03 & 0.02 & 0.10 & 0.16 & 0.22 & -0.04 & 0.11 & 0.18 & 0.16 & -0.01 \\ 
\hline\hline
\multicolumn{13}{c}{\textbf{Multi-Ref LORs}} \\ \hline
\multicolumn{1}{c|}{\multirow{2}{*}{\textbf{Embedding}}} & \multicolumn{4}{c|}{\textbf{w/o Redundancy}} & \multicolumn{4}{c}{\textbf{+ Redundancy Penalty (ROUGE)}} & \multicolumn{4}{|c}{\textbf{+ Redundancy Penalty (BERTScore)}} \\ \cline{2-13} 
\multicolumn{1}{c|}{} & \multicolumn{1}{l}{Consistency} & \multicolumn{1}{l}{Relevance} & \multicolumn{1}{l}{Coherence} & \multicolumn{1}{l}{Fluency} & \multicolumn{1}{|l}{Consistency} & \multicolumn{1}{l}{Relevance} & \multicolumn{1}{l}{Coherence} & \multicolumn{1}{l}{Fluency} & \multicolumn{1}{|l}{Consistency} & \multicolumn{1}{l}{Relevance} & \multicolumn{1}{l}{Coherence} & \multicolumn{1}{l}{Fluency}\\ \hline
Infersent & 0.03 & 0.09 & 0.07 & \cellcolor{green!15}{\textbf{0.08}} & 0.05 & 0.15 & \cellcolor{green!15}{\textbf{0.23}} & 0.04 & 0.02 & 0.14 & 0.16 & 0.05 \\
Elmo & 0.03 & 0.04 & -0.04 & 0.03 & 0.04 & 0.10 & 0.14 & -0.03 & 0.03 & 0.09 & 0.06 & -0.01 \\
STSb-bert & 0.09 & 0.10 & 0.00 & 0.01 & 0.10 & \cellcolor{green!15}{\textbf{0.16}} & 0.19 & -0.06 & 0.09 & 0.17 & 0.13 & -0.03 \\
STSb-roberta & 0.10 & 0.05 & -0.06 & 0.00 & 0.11 & 0.13 & 0.15 & -0.08 & 0.09 & 0.12 & 0.07 & -0.04 \\
USE & 0.04 & 0.08 & 0.03 & 0.04 & 0.05 & 0.14 & 0.22 & -0.04 & 0.03 & 0.13 & 0.15 & 0.01 \\
STSb-distilbert & \cellcolor{green!15}{\textbf{0.13}} & 0.06 & 0.01 & -0.01 & 0.12 & 0.11 & 0.17 & -0.09 & 0.12 & 0.12 & 0.12 & -0.06 \\ \hline
\multicolumn{13}{c}{\textbf{Multi-Ref MORs}} \\ \hline
\multicolumn{1}{c|}{\multirow{2}{*}{\textbf{Embedding}}} & \multicolumn{4}{c|}{\textbf{w/o Redundancy}} & \multicolumn{4}{c}{\textbf{+ Redundancy Penalty (ROUGE)}} & \multicolumn{4}{|c}{\textbf{+ Redundancy Penalty (BERTScore)}} \\ \cline{2-13} 
\multicolumn{1}{c|}{} & \multicolumn{1}{l}{Consistency} & \multicolumn{1}{l}{Relevance} & \multicolumn{1}{l}{Coherence} & \multicolumn{1}{l}{Fluency} & \multicolumn{1}{|l}{Consistency} & \multicolumn{1}{l}{Relevance} & \multicolumn{1}{l}{Coherence} & \multicolumn{1}{l}{Fluency} & \multicolumn{1}{|l}{Consistency} & \multicolumn{1}{l}{Relevance} & \multicolumn{1}{l}{Coherence} & \multicolumn{1}{l}{Fluency}\\ \hline
Infersent & 0.06 & 0.10 & 0.05 & 0.06 & 0.07 & 0.19 & \cellcolor{green!15}{\textbf{0.26}} & -0.01 & \cellcolor{green!15}{\textbf{0.06}} & 0.18 & \cellcolor{green!15}{\textbf{0.19}} & 0.02 \\
Elmo & 0.06 & 0.06 & 0.00 & 0.02 & 0.04 & 0.13 & 0.17 & -0.06 & 0.04 & 0.12 & 0.11 & -0.02 \\
STSb-bert & 0.08 & 0.01 & -0.02 & 0.01 & \cellcolor{green!15}{\textbf{0.09}} & 0.09 & 0.18 & -0.08 & 0.08 & 0.08 & 0.11 & -0.04 \\
STSb-roberta & 0.05 & 0.07 & 0.00 & 0.02 & 0.06 & 0.14 & 0.20 & -0.07 & 0.05 & 0.14 & 0.13 & -0.02 \\
USE & 0.01 & 0.09 & 0.05 & 0.01 & 0.04 & 0.16 & 0.24 & -0.05 & 0.01 & 0.16 & 0.19 & -0.02 \\
STSb-distilbert & 0.08 & 0.02 & 0.00 & -0.01 & 0.07 & 0.09 & 0.18 & -0.09 & 0.07 & 0.08 & 0.12 & -0.06 \\ \hline\hline
\multicolumn{13}{c}{\textbf{Multi-Ref HORs}} \\ \hline
\multicolumn{1}{c|}{\multirow{2}{*}{\textbf{Embedding}}} & \multicolumn{4}{c|}{\textbf{w/o Redundancy}} & \multicolumn{4}{c}{\textbf{+ Redundancy Penalty (ROUGE)}} & \multicolumn{4}{|c}{\textbf{+ Redundancy Penalty (BERTScore)}} \\ \cline{2-13} 
\multicolumn{1}{c|}{} & \multicolumn{1}{l}{Consistency} & \multicolumn{1}{l}{Relevance} & \multicolumn{1}{l}{Coherence} & \multicolumn{1}{l}{Fluency} & \multicolumn{1}{|l}{Consistency} & \multicolumn{1}{l}{Relevance} & \multicolumn{1}{l}{Coherence} & \multicolumn{1}{l}{Fluency} & \multicolumn{1}{|l}{Consistency} & \multicolumn{1}{l}{Relevance} & \multicolumn{1}{l}{Coherence} & \multicolumn{1}{l}{Fluency}\\ \hline
Infersent & 0.09 & 0.11 & 0.06 & 0.05 & 0.13 & 0.18 & 0.25 & -0.01 & 0.09 & 0.18 & 0.18 & 0.02 \\
Elmo & 0.05 & 0.08 & 0.02 & 0.05 & 0.07 & 0.16 & 0.19 & -0.01 & 0.05 & 0.14 & 0.12 & 0.02 \\
STSb-bert & 0.07 & 0.11 & 0.04 & 0.05 & 0.11 & 0.18 & 0.25 & -0.02 & 0.06 & 0.19 & 0.17 & 0.02 \\
STSb-roberta & 0.10 & 0.08 & 0.01 & 0.04 & \cellcolor{green!15}{\textbf{0.13}} & 0.16 & 0.21 & -0.04 & 0.11 & 0.15 & 0.13 & 0.00 \\
USE & 0.06 & 0.13 & 0.07 & \cellcolor{green!15}{\textbf{0.05}} & 0.09 & \cellcolor{green!15}{\textbf{0.20}} & \cellcolor{green!15}{\textbf{0.26}} & -0.02 & 0.06 & 0.20 & 0.19 & 0.02 \\
STSb-distilbert & 0.09 & 0.09 & 0.03 & 0.03 & 0.12 & 0.15 & 0.22 & -0.05 & 0.10 & 0.15 & 0.15 & 0.00 \\ \hline
\end{tabular}}
\caption{Kendall Tau ($\tau$) correlation coefficient for \textbf{Ensemble\textsubscript{rel}} when lambda ($\lambda)$ $= 0.5$ for consistency, relevance, coherence and fluency dimension without redundancy and when ROUGE and BERTScore is used as redundancy penalty for different terminology variations of multiple references (highly abstractive (LORs), medium overlapping (MORs) and highly extractive (HORs) references). The best value in each dimension has been bold green.}
\label{tab:hum_multi_ref_en_rel}
\end{table*}

\section{Related Work}
The most common method for evaluating model summaries has been to compare them against human-written reference summaries. ROUGE~\cite{lin-2004-rouge} considers direct lexical overlap and afterwards different version of ROUGE~\cite{DBLP:conf/emnlp/Graham15} has also been proposed including \textit{ROUGE} with word embedding~\cite{DBLP:conf/emnlp/NgA15} and synonym~\cite{DBLP:journals/corr/abs-1803-01937}, graph-based lexical measurement~\cite{DBLP:conf/emnlp/ShafieiBavaniEW18}, Vanilla \textit{ROUGE}~\cite{DBLP:conf/acl/YangLLLL18} and highlight-based \textit{ROUGE}~\cite{DBLP:conf/acl/HardyNV19} to mitigate the limitations of original ROUGE. Metrics based on semantic similarity between reference and model summaries have also been proposed to capture the semantics, including  S+WMS~\cite{DBLP:conf/acl/ClarkCS19}, MoverScore~\cite{DBLP:conf/emnlp/ZhaoPLGME19}, and BERTScore~\cite{DBLP:conf/iclr/ZhangKWWA20}. Reference-free evaluation has also been a recent trend to avoid dependency on human reference~\cite{DBLP:conf/emnlp/BohmGMSDG19, DBLP:conf/acl/Peyrard19, DBLP:conf/emnlp/SunN19, DBLP:conf/acl/GaoZE20, DBLP:conf/emnlp/WuMWMJ20}.

%In fact, the widely used method underlying \textit{extractive} summarization involves ranking original document sentences according to how well they capture the overall description, then concatenating the top-ranked sentences to form a summary. Thus, the ``right'' evaluation metric for the \textit{extractive} summarization task should also consider the quality of the sentence ranker. 

Although the \textit{extractive} summarizing task is typically framed as a sentence ranking problem, none of the mentioned metrics evaluate the quality of the ranker. To address this, recently~\cite{DBLP:conf/acl/AkterBS22} has proposed a rank-aware and gain-based evaluation metric for extractive summarization called \textit{Sem-nCG}, but it does not incorporate redundancy and also lacks evaluation with multiple references. These are two significant limitations that need to be addressed, and hence, the focus of this work.

Redundancy in extracted sentences is a prominent issue in extractive summarization systems. Maximal Marginal Relevance (MMR)~\cite{DBLP:conf/sigir/CarbonellG98} is a classic algorithm to penalize redundancy in model summary. There are several approaches that explicitly model redundancy and use algorithms to avoid selecting sentences that are too similar to those that have already been extracted~\cite{DBLP:conf/coling/RenWCMZ16}. Trigram blocking~\cite{DBLP:conf/iclr/PaulusXS18} is another popular approach to reduce redundancy in model summary.~\citet{DBLP:conf/acl/ChenLK20} has shown how to compute self-referenced redundancy score while evaluating the model summary. 

% In the context of multi-reference summary evaluation, our work is additionally distinctive since we do not follow the conventional procedure of computing the evaluation metric for each reference separately and then estimating their average/max. Instead, we use a variety of human-written reference summaries to infer a single, unified ground-truth ranked list of source sentences, after which the sem-nCG score is computed only once.

When multiple reference summaries are available, Researchers have also suggested Pyramid-based~\cite{DBLP:conf/naacl/NenkovaP04} approaches for summary evaluation. However, this method requires more manual labor and has undergone numerous improvements~\cite{DBLP:conf/acl/PassonneauCGP13, DBLP:conf/aaai/YangPM16, DBLP:conf/naacl/ShapiraG0RPBAD19, DBLP:conf/acl/MaoLZRH20}, it still needs a substantial amount of manual effort, making it unsuitable for large-scale evaluation. Recently, for NLG evaluation different unified frameworks and models~\cite{DBLP:conf/emnlp/DengTLXH21, DBLP:journals/corr/abs-2210-07197, DBLP:conf/emnlp/LiuIXWXZ23, DBLP:conf/eacl/WuIPBH24} to predict different aspects of the generated text has been proposed. Even though these metrics can be applied to text summarization, it is still a data-driven approach and it is unclear why the model produces such scores.

\underline{Uniqueness to our work:} We improved the sem-nCG metric for extractive summarization, in order to make it more aware of redundancy. This was tricky, as it requires a balance of importance and diversity during evaluation. We also showed how to use the updated metric for multiple references, which was challenging due to variations in human references and terminology.

\section{Conclusion}
Previous work has proposed the \textit{Sem-nCG} metric exclusively for evaluating extractive summarization task considering both rank awareness and semantics. However, the \textit{Sem-nCG} metric ignores redundancy in a model summary and does not support evaluation with multiple reference summaries, which are two significant limitations. In this paper, we proposed a redundancy-aware multi-reference based \textit{Sem-nCG} metric which is superior compared to the previous \textit{Sem-nCG} metric along \textit{Consistency}, \textit{Relevance} and \textit{Coherence} dimensions. Additionally, for summary evaluation using multiple references, we created a unique ground-truth ranking by incorporating multiple references rather than trivial max/average score computation with multiple references. Our empirical evaluation shows that the traditional metric becomes unstable when multiple references are available and the revised redundancy-aware \textit{Sem-nCG} shows a notably higher correlation with human judgments than ROUGE and BERTScore metric both for single and multiple references. Thus we encourage the community to evaluate extractive summaries using the revised redundancy-aware \textit{Sem-nCG} metric.

\section{Limitations}
One limitation of the work is that the dataset for human evaluation is not big (252 samples). We used the dataset from~\cite{DBLP:journals/tacl/FabbriKMXSR21}, the only available benchmark "meta-evaluation dataset" for extractive summarization, to the best of our knowledge.~\cite{DBLP:conf/acl/AkterBS22} have demonstrated the correlation of \textit{Sem-nCG} with human judgment on this dataset. To ensure a fair comparison, we maintained the same settings as the original \textit{Sem-nCG} when assessing the redundancy-aware \textit{Sem-nCG}. To evaluate the redundancy-aware \textit{Sem-nCG} we will need a similar kind of evaluation benchmark and we can not do anything here. Even though~\cite{liu-etal-2023-revisiting} has published a new dataset, that work focuses mainly on how to increase human annotation reliability for summary evaluation with respect to Atomic Content Unit (ACU) and doesn’t provide human judgment for model’s summary along four summary quality dimensions: coherence, consistency, fluency and relevance.

Another limitation of the work may seem like that the ablation study does not show any consistent pattern. We understand that it's difficult to come up with a single evaluation metric that can account for different qualities such as coherence, consistency, fluency, and relevance. It requires careful consideration to balance these different qualities. However, we noticed that extractive sentences are inherently grammatically correct, so we can exclude fluency from the hyperparameter choice. After analyzing the data, we found that a balanced $\lambda$ value of 0.5 worked well across all four quality dimensions. This suggests that this configuration strikes a reasonable tradeoff between importance and diversity. It addresses the complexities inherent in assessing summarization quality comprehensively with a single score from the metric.

\section{Ethics Statement}
For the experiments, we used a publicly accessible dataset and anonymous human annotations. As a result, to the best of our knowledge, there are no ethical violations. Additionally, the evaluation of extractive summarization is a major aspect of this work. Hence, we consider it a low-risk research study.

\section{Acknowledgements}
This work has been partially supported by the Research Center Trustworthy Data Science and Security \url{https://rc-trust.ai}, one of the
Research Alliance centers within the \url{https://uaruhr.de}. This work has also received partial support from the National Science Foundation (NSF) Standard Grant Award \#2302974 and the Air Force Office of Scientific Research Grant/Cooperative Agreement Award \#FA9550-23-1-0426.

\bibliography{references}

\appendix
\section{Appendix}
\subsection{Explanation of Metrics for \textit{Score\textsubscript{red}}} \label{metric}
\noindent{\bf {ROUGE~\cite{lin-2004-rouge}: }} Between the generated summary and reference summary, ROUGE counts the overlap of textual units (n-grams, word sequences).

\noindent{\bf {MoverScore~\cite{DBLP:conf/emnlp/ZhaoPLGME19}:}} uses the Word Mover's Distance~\cite{DBLP:conf/icml/KusnerSKW15} to calculate the semantic distance between a summary and a reference text, pooling n-gram embedding from BERT representations.

\noindent{\bf {BERTScore~\cite{DBLP:conf/iclr/ZhangKWWA20}:}} calculates similarity scores by matching generated and reference summaries on a token level. The cosine similarity between contextualized token embeddings from BERT is maximized by computing token matching greedily.

\noindent{\bf {Cosine Similarity:}} Sentences are converted to sentence embedding using STSb-distilbert~\cite{reimers-2019-sentence-bert}. Then the semantic similarity of sentences is measured using cosine similarity between sentence vectors.

The code for the metrics used can be found here.\footnote{\url{https://github.com/Yale-LILY/SummEval/tree/master/evaluation/summ_eval}}

\subsection{Human Evaluation Components} \label{hum_dimension}
To calculate the Kendall's Tau ($\tau$) rank correlation for the redundancy-aware \textit{Sem-nCG} metric, we used four quality dimensions following~\cite{DBLP:conf/acl/AkterBS22, DBLP:journals/tacl/FabbriKMXSR21}.

\noindent{\bf{Consistency}:} refers to the fact that the contents in the summary and the source are the same. Only assertions from the source are included in factually consistent summaries, which do not include any trippy facts.

\noindent{\bf{Relevance}:} getting the most important information from a source. The annotators were to penalize summaries with redundancy and excessive information. In the summary, only important information from the source should be included. 

\noindent{\bf{Coherence}:} overall summary sentence quality while keeping a coherent body of information on a topic rather than a tangle of related information~\cite{dang2005overview}.

\noindent{\bf{Fluency}:} the structure and quality of the summary sentences. As mentioned in~\cite{dang2005overview} ``should have no formatting problems, capitalization errors or obviously ungrammatical sentences (e.g., fragments, missing components) that make the text difficult to read.''

\subsection{Computational Infrastructure \& Runtime} \label{comp}
\begin{table}[!htb]
\centering
\begin{tabular}{llc}
\hline
\multicolumn{3}{c}{\textbf{Computational Infrastructure}} \\ 
\multicolumn{3}{c}{NVIDIA Quadro RTX 5000 GPUs} \\ \hline
\multicolumn{3}{c}{\textbf{Hyperparameter Search}} \\ 
\multicolumn{3}{c}{$\lambda \in [0, 1]$ uniform-integer distribution}\\ \hline
\multicolumn{1}{c|}{\textbf{Type}} & \multicolumn{1}{c|}{\textbf{Variation}} & \textbf{Runtime (s)} \\ \hline
\multicolumn{1}{l|}{\multirow{4}{*}{\textit{Score\textsubscript{red}}}} & \multicolumn{1}{l|}{Cosine Similarity} &  0.06 \\ 
\multicolumn{1}{l|}{} & \multicolumn{1}{l|}{ROUGE} & 0.44 \\ 
\multicolumn{1}{l|}{} & \multicolumn{1}{l|}{MoverScore} &  0.23 \\  
\multicolumn{1}{l|}{} & \multicolumn{1}{l|}{BERTScore} & 14.7 \\ \hline
\multicolumn{1}{l|}{\multirow{7}{*}{\textit{Sem-nCG}}} & \multicolumn{1}{l|}{Infersent} & 0.4 \\ 
\multicolumn{1}{l|}{} & \multicolumn{1}{l|}{Elmo} & 79.1 \\ 
\multicolumn{1}{l|}{} & \multicolumn{1}{l|}{STSb-bert} & 0.33 \\ 
\multicolumn{1}{l|}{} & \multicolumn{1}{l|}{STSb-roberta} & 0.34 \\  
\multicolumn{1}{l|}{} & \multicolumn{1}{l|}{USE} & 20.2 \\  
\multicolumn{1}{l|}{} & \multicolumn{1}{l|}{STSb-distilbert} & 0.13 \\ 
\multicolumn{1}{l|}{} & \multicolumn{1}{l|}{Ensemble\textsubscript{sim}} & 20.33 \\ \hline
\end{tabular}
\end{table}

% \subsection{Kendall's Tau ($\tau$) Correlation for ROUGE and BERTScore}
% Comparing the current redundancy-aware Sem-nCG metric with alternative metrics is orthogonal goal of this paper. However, Kendall's tau ($\tau$) correlation with human given score of model summary in \textit{Consistency, Relevance, Coherence, and Fluency} dimensions for ROUGE and BERTScore is represented by Table~\ref{tab:hum_rouge_bert}. 

\subsection{Sentence Embedding Used in Setion~\ref{result}}
\noindent{\bf{Infersent}~\cite{DBLP:conf/emnlp/ConneauKSBB17}:} 
Infersent-v2 is trained with fastText word embedding and generates 4096-dimensional sentence embedding using a BiLSTM network with max-pooling.

\noindent{\bf{Elmo}~\cite{DBLP:conf/naacl/PetersNIGCLZ18}:}
The contextualized word embedding was transformed into a sentence embedding using a fixed mean-pooling of all contextualized word representations with embedding shape 1024.

\noindent{\bf{Google Universal Sentence Encoder (USE)}~\cite{cer-etal-2018-universal}:} We utilized USE with enc-2~\cite{iyyer-etal-2015-deep} which is based on the deep average network to transform input text to a 512-dimensional sentence embedding.

\noindent{\bf{Semantic Textual Similarity benchmark (STSb)}~\cite{reimers-2019-sentence-bert}:} Sentence Transformer allows to generate dense vector
representations of sentences. Three of the best available models that were optimized for semantic textual similarity were considered: STSb-bert (embedding size 1024),
STSb-roberta (embedding size 1024) and STSb-distilbert (embedding size 768).

% graph
\setlength{\textfloatsep}{0pt}
\begin{figure*}[!htb]
        \begin{subfigure}[b]{\textwidth}
              \centering
              \frame{\includegraphics[width=\linewidth,trim={7 280 10 240},clip]{images/bertscore/legend.pdf}}
            \vspace{-4.5 mm}
        \end{subfigure}
        \hfill%
        \begin{subfigure}[b]{0.32\textwidth}
              \includegraphics[width=\linewidth,trim={10 10 155 100},clip]{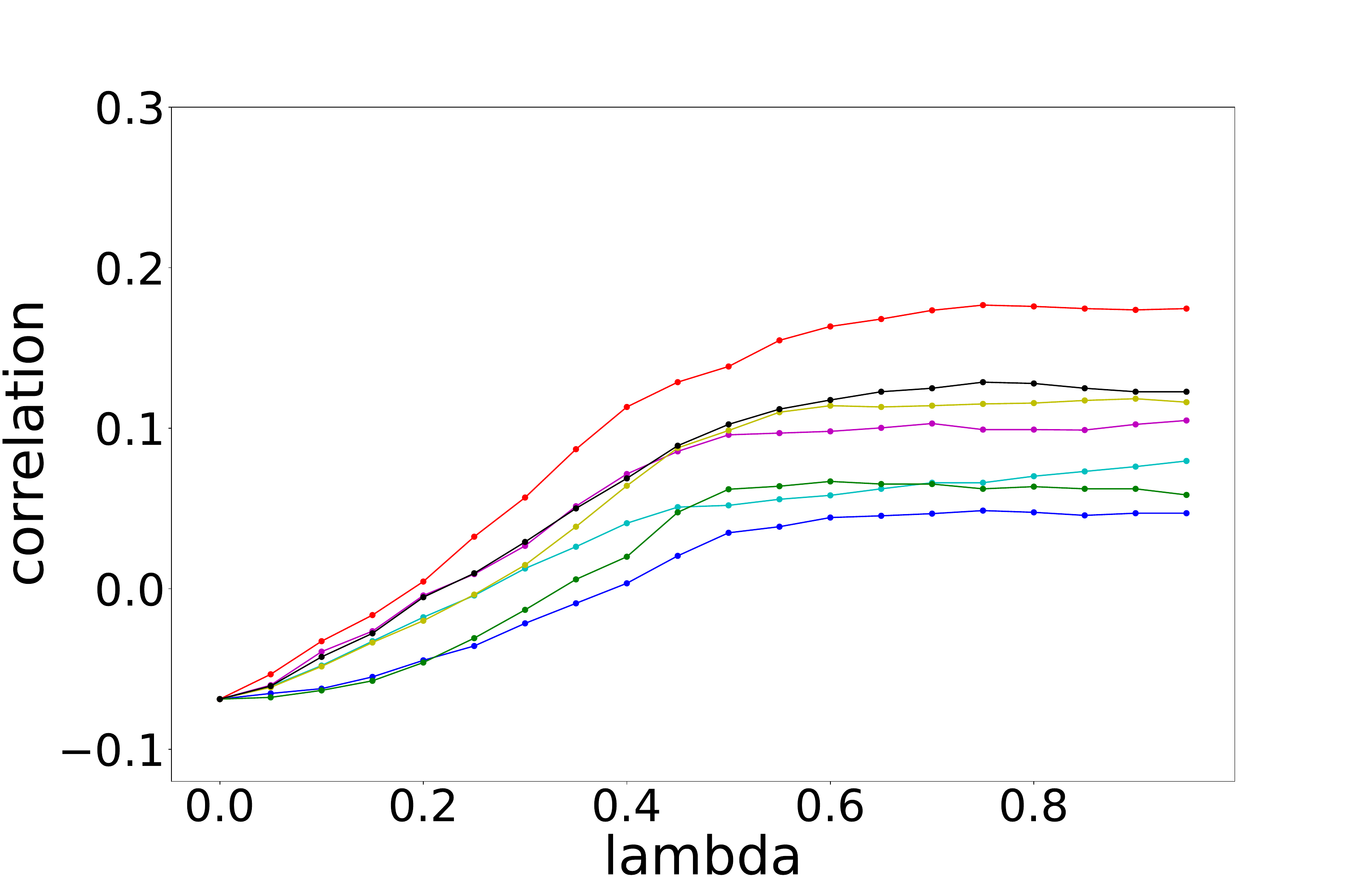}
              \caption{LOR - Consistency}
        \end{subfigure}
        \hfill%
        \begin{subfigure}[b]{0.32\textwidth}
              \includegraphics[width=\linewidth,trim={10 10 155 100},clip]{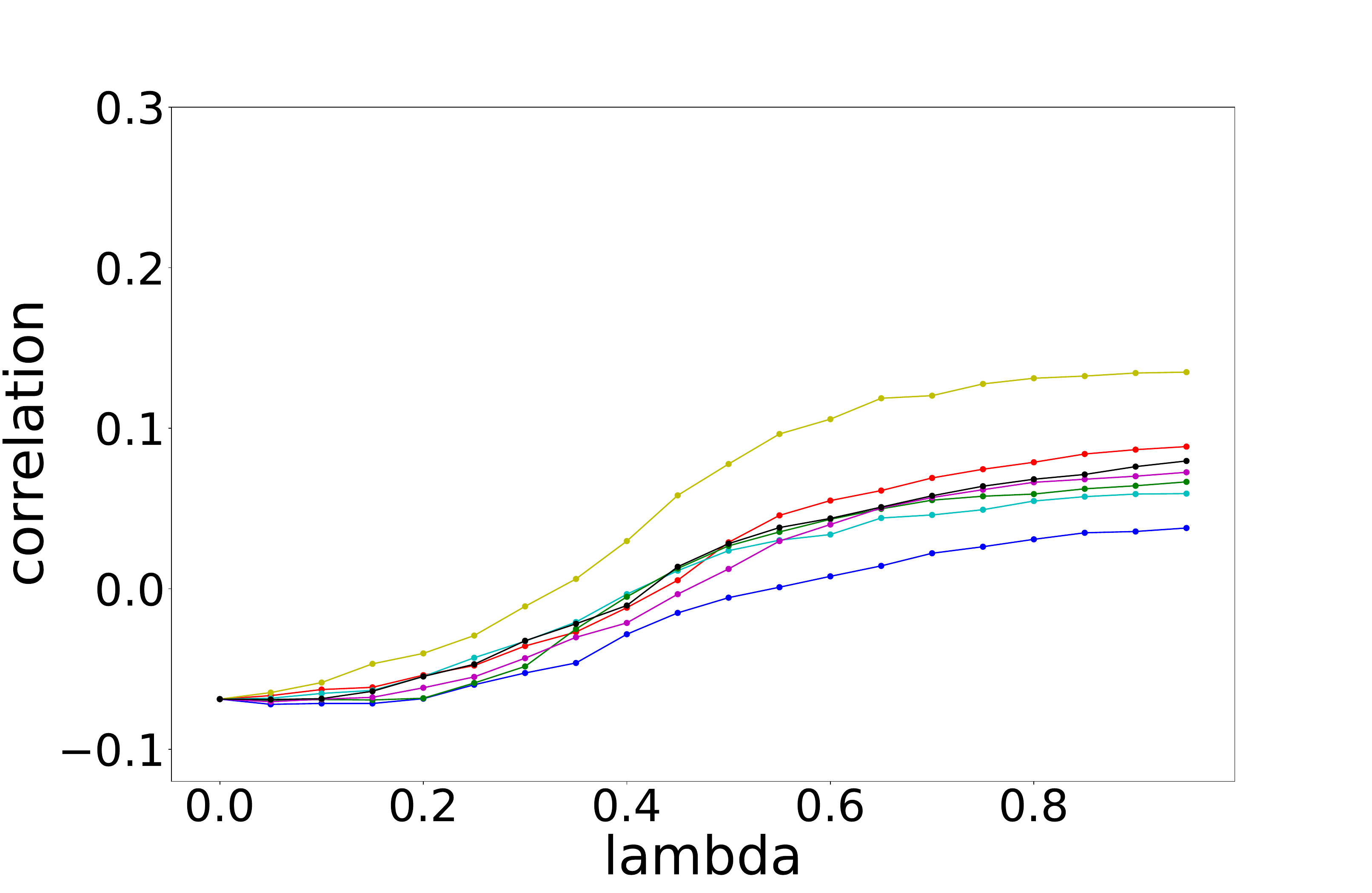}
            \caption{MOR - Consistency}
        \end{subfigure}
        \hfill%
        \begin{subfigure}[b]{0.32\textwidth}
              \includegraphics[width=\linewidth,trim={10 10 155 100},clip]{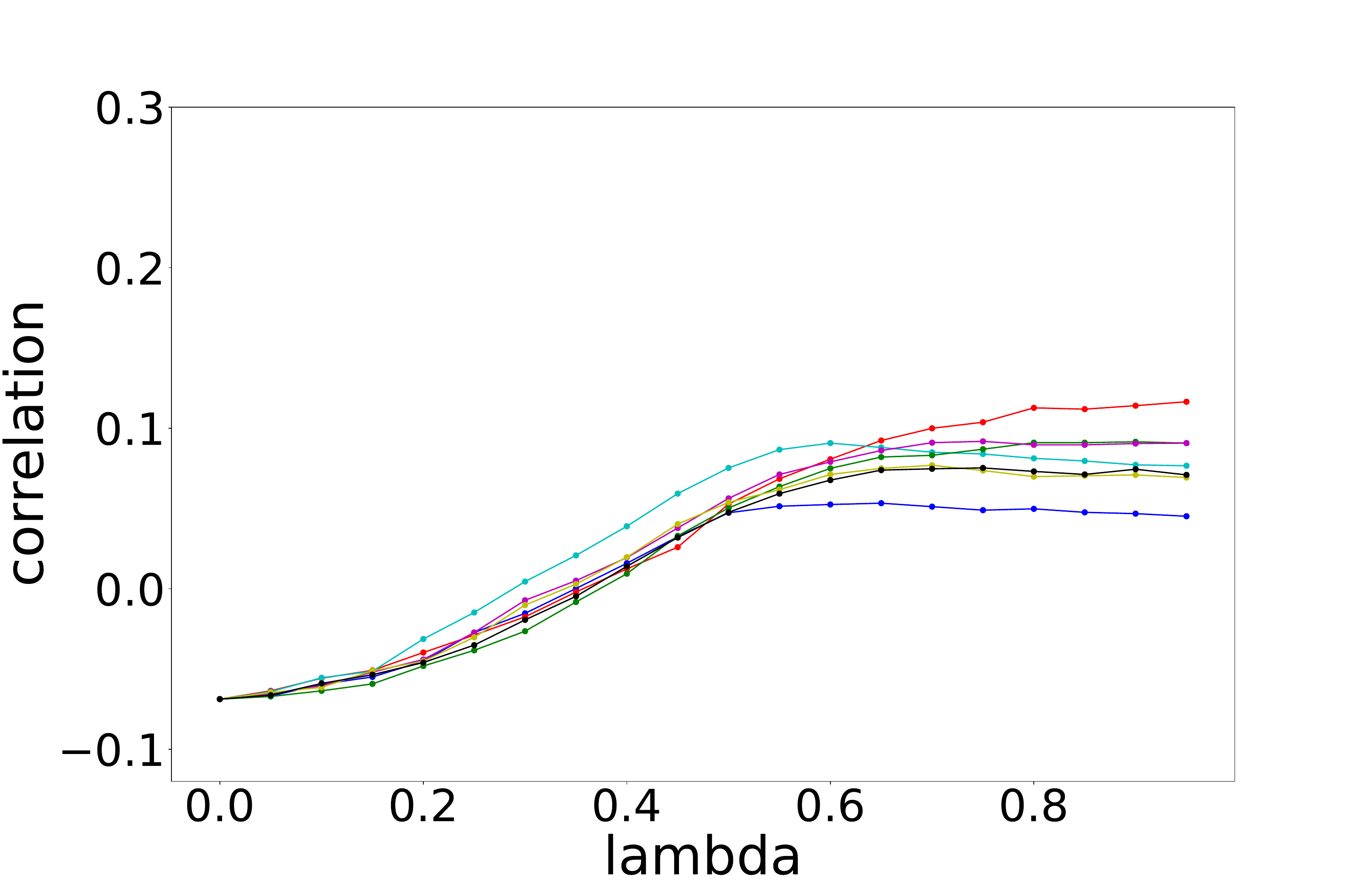}
              \caption{HOR - Consistency}
        \end{subfigure}
        
        \begin{subfigure}[b]{0.32\textwidth}
              \includegraphics[width=\linewidth,trim={10 10 155 105},clip]{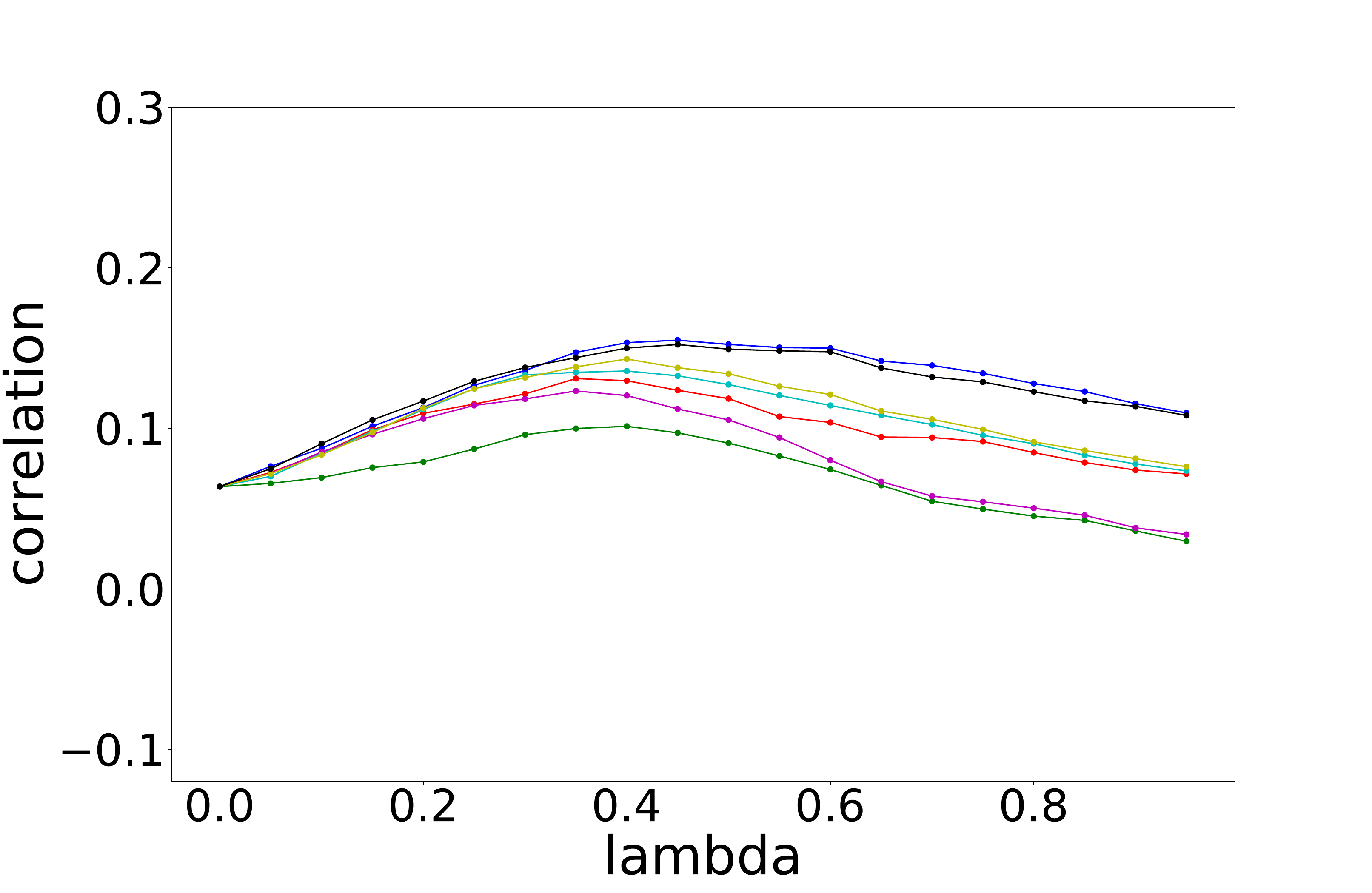}
              \caption{LOR - Relevance}
        \end{subfigure}
        \hfill%
        \begin{subfigure}[b]{0.32\textwidth}
              \includegraphics[width=\linewidth,trim={10 10 155 105},clip]{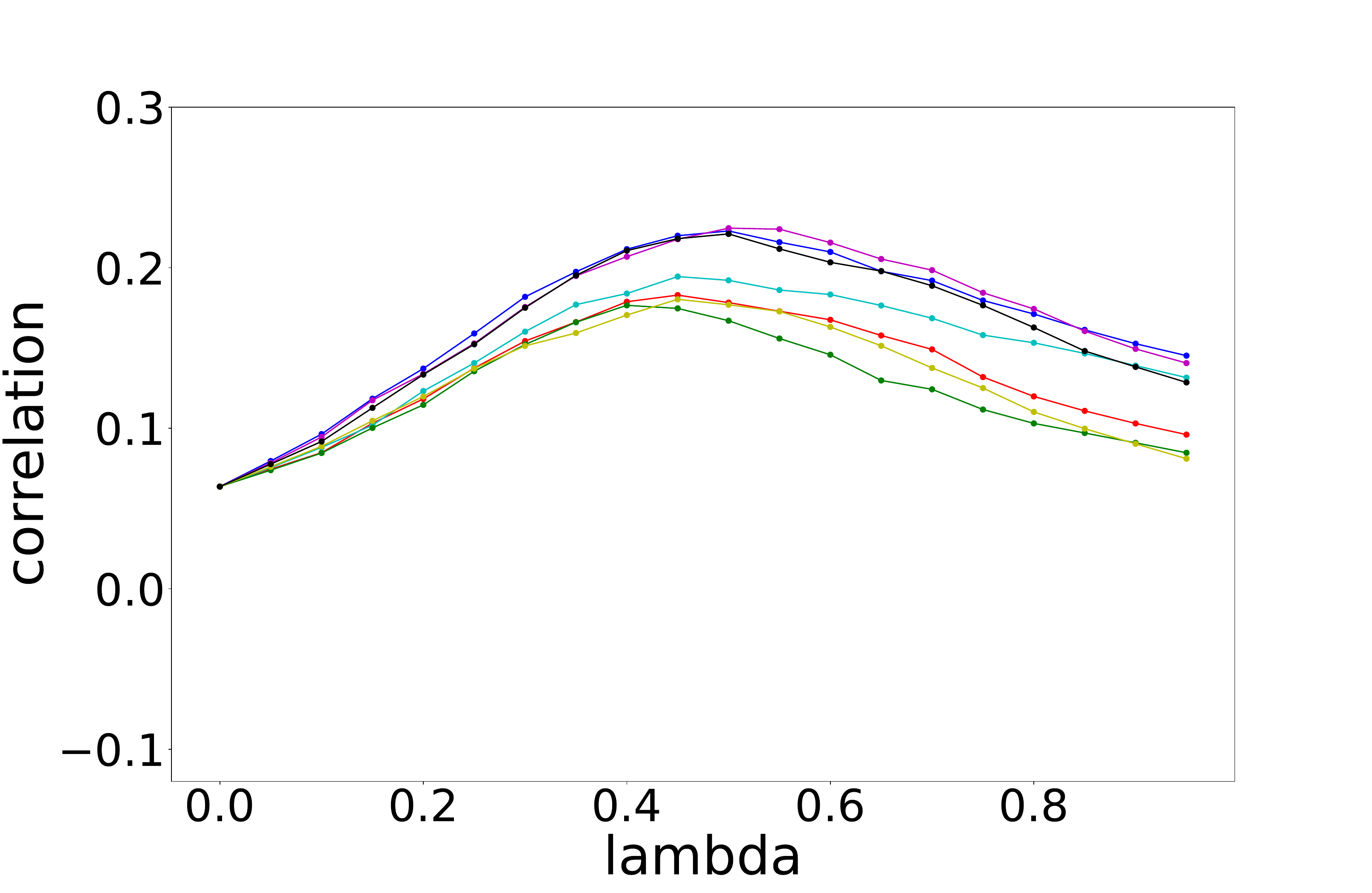}
            \caption{MOR - Relevance}
        \end{subfigure}
        \hfill%
        \begin{subfigure}[b]{0.32\textwidth}
              \includegraphics[width=\linewidth,trim={10 10 155 105},clip]{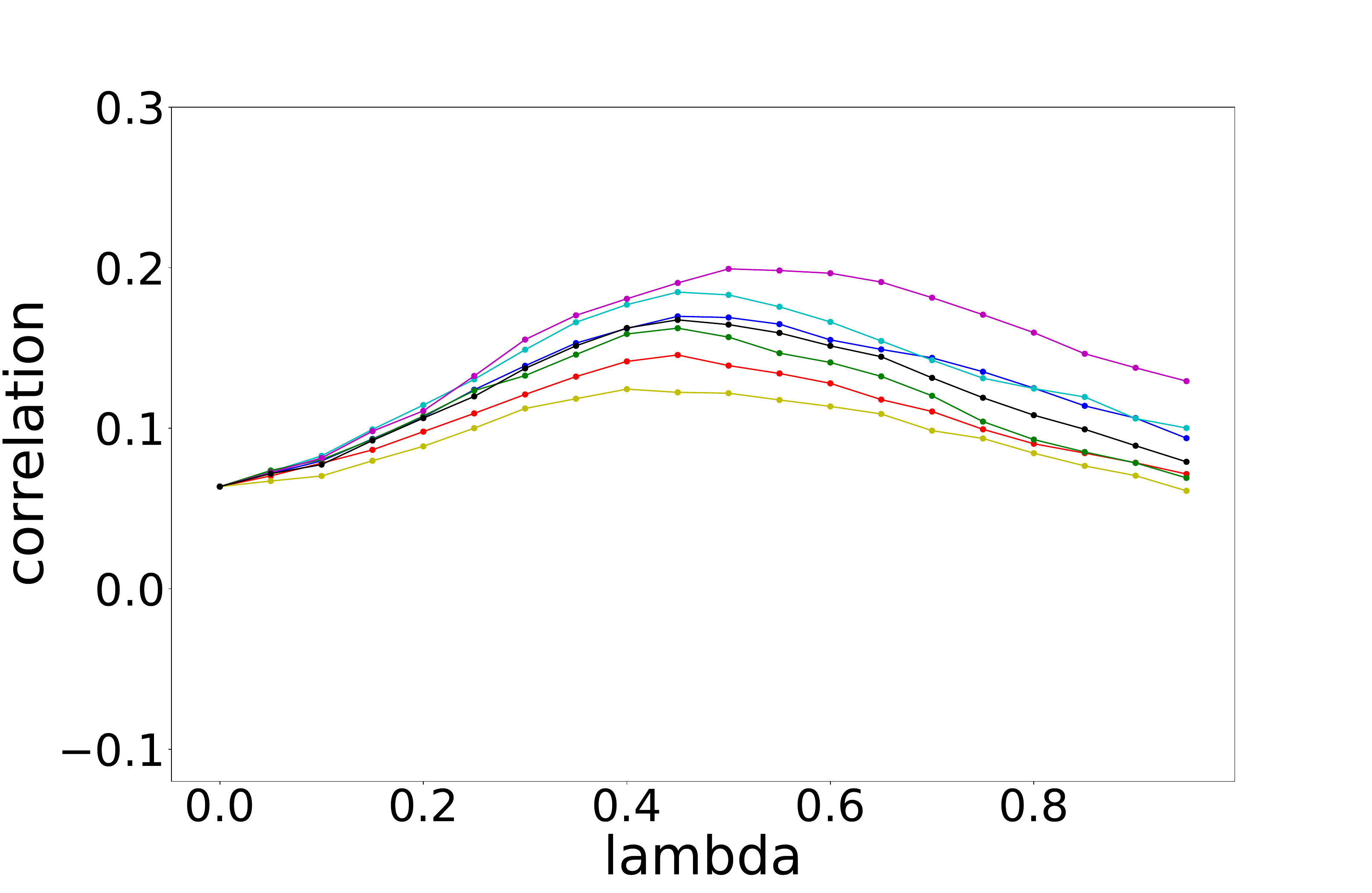}
              \caption{HOR - Relevance}
        \end{subfigure}
        
        \begin{subfigure}[b]{0.32\textwidth}
              \includegraphics[width=\linewidth,trim={10 10 155 105},clip]{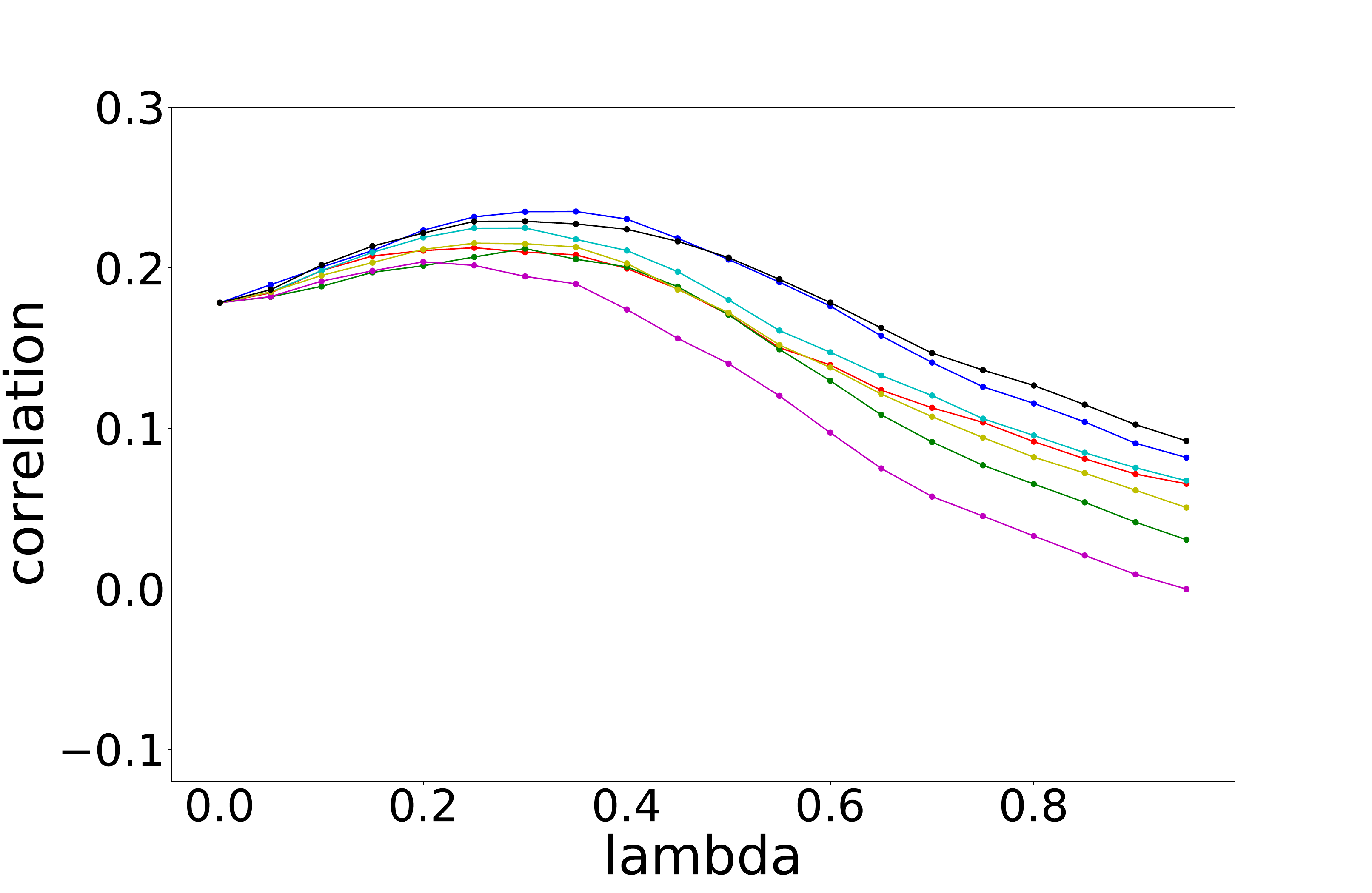}
              \caption{LOR - Coherence}
        \end{subfigure}
        \hfill%
        \begin{subfigure}[b]{0.32\textwidth}
              \includegraphics[width=\linewidth,trim={10 10 155 105},clip]{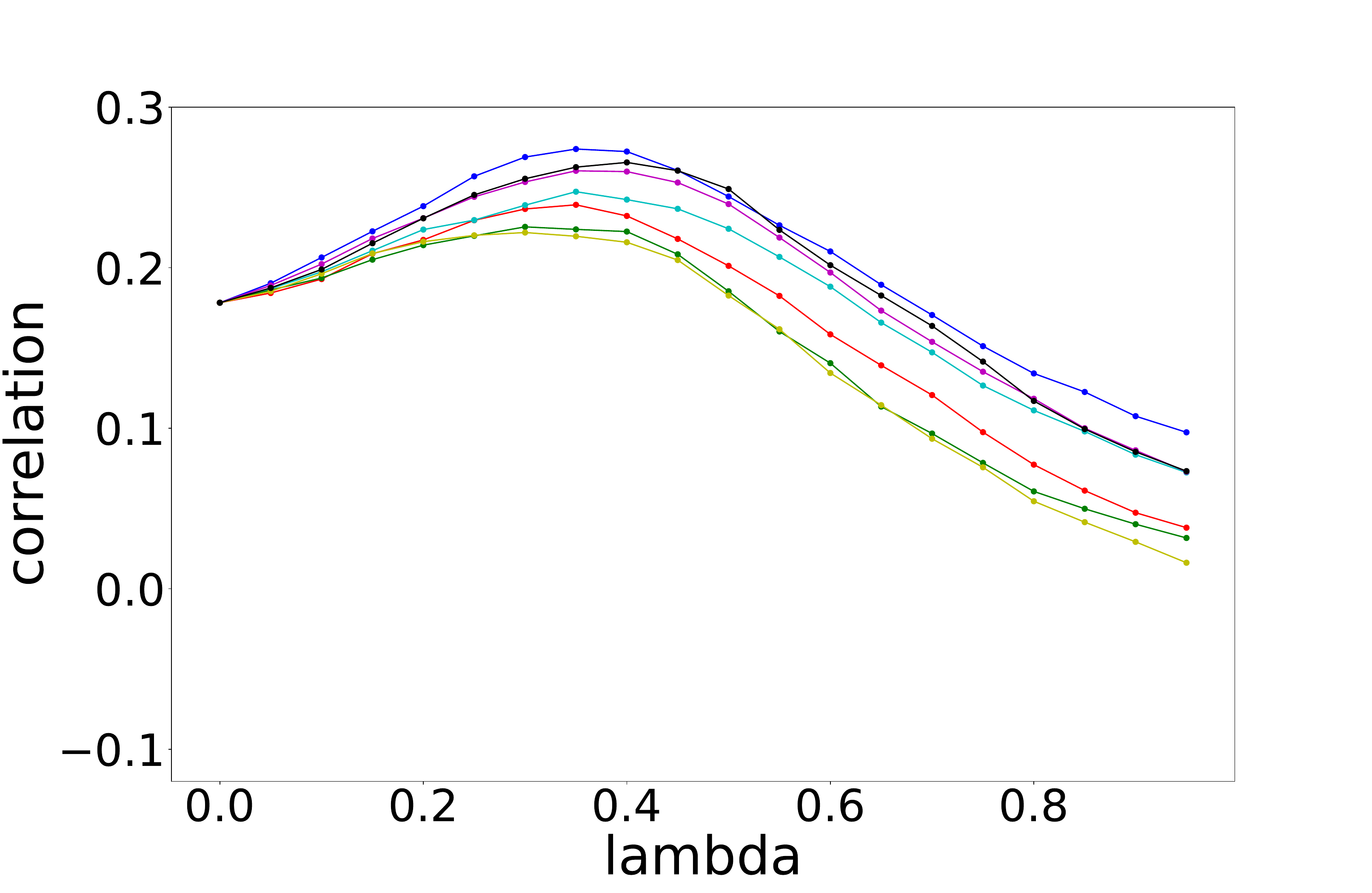}
            \caption{MOR - Coherence}
        \end{subfigure}
        \hfill%
        \begin{subfigure}[b]{0.32\textwidth}
              \includegraphics[width=\linewidth,trim={10 10 155 105},clip]{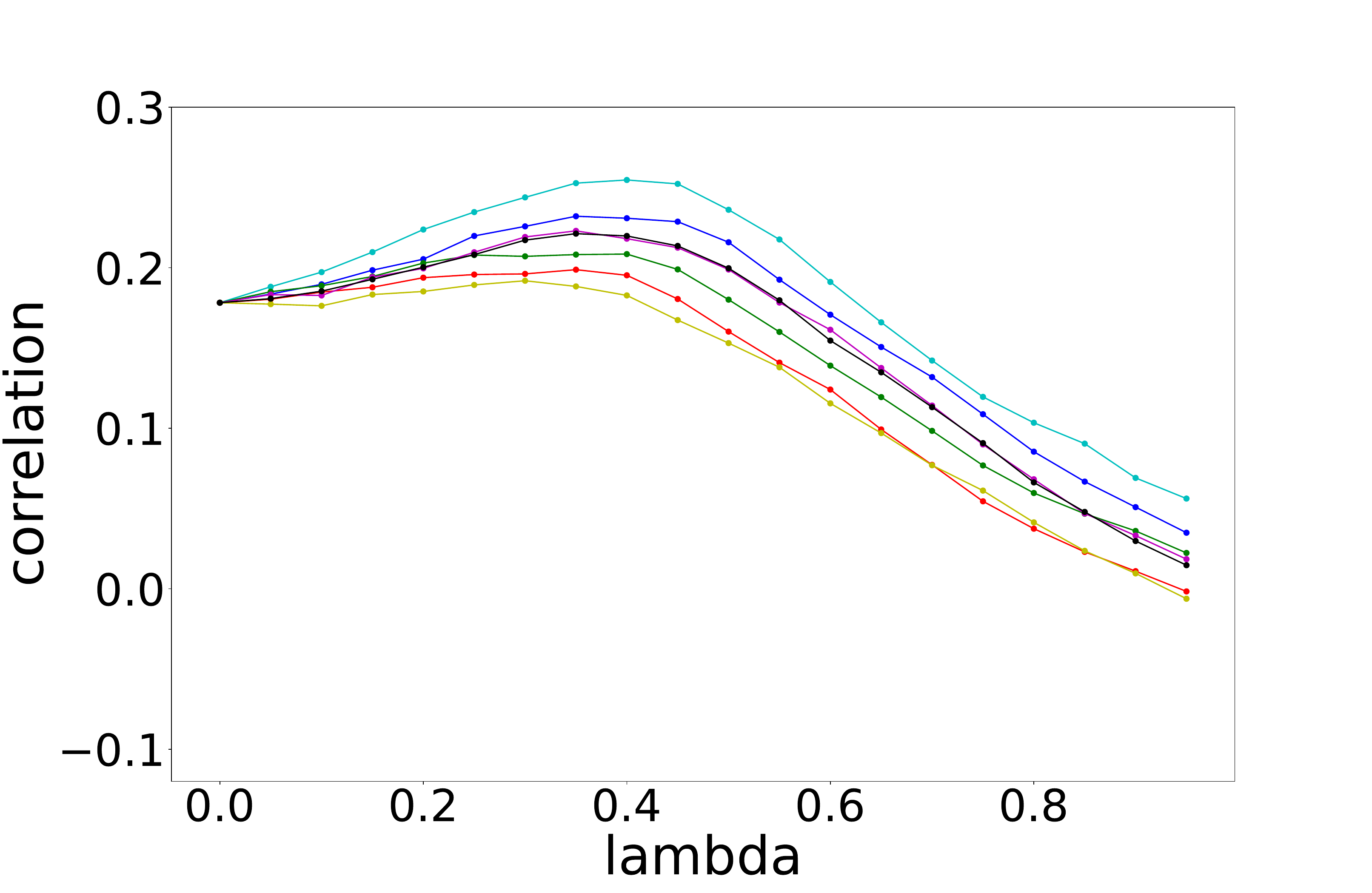}
              \caption{HOR - Coherence}
        \end{subfigure}
        
        \begin{subfigure}[b]{0.32\textwidth}
              \includegraphics[width=\linewidth,trim={10 10 155 105},clip]{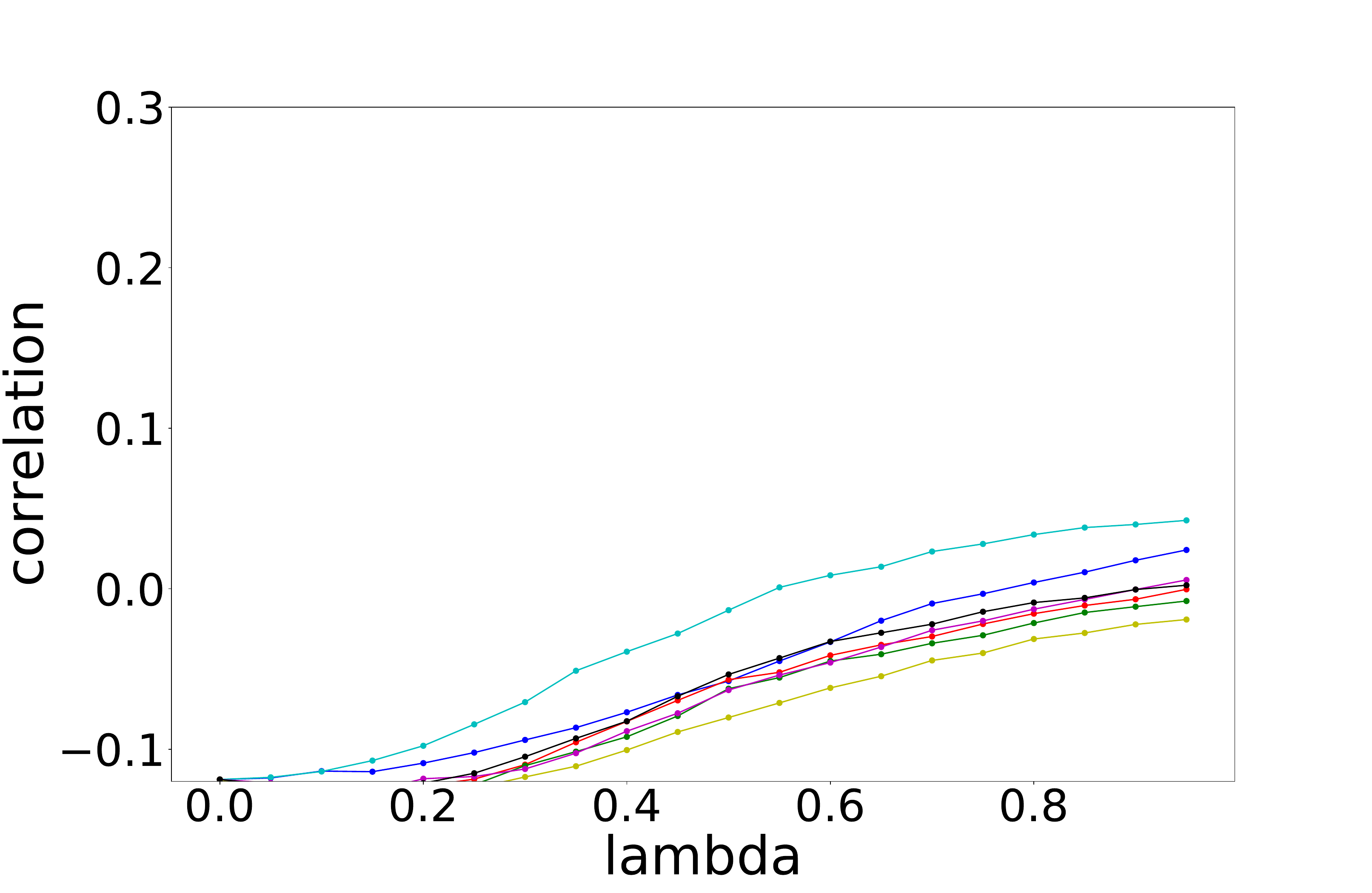}
              \caption{LOR - Fluency}
        \end{subfigure}
        \hfill%
        \begin{subfigure}[b]{0.32\textwidth}
              \includegraphics[width=\linewidth,trim={10 10 155 105},clip]{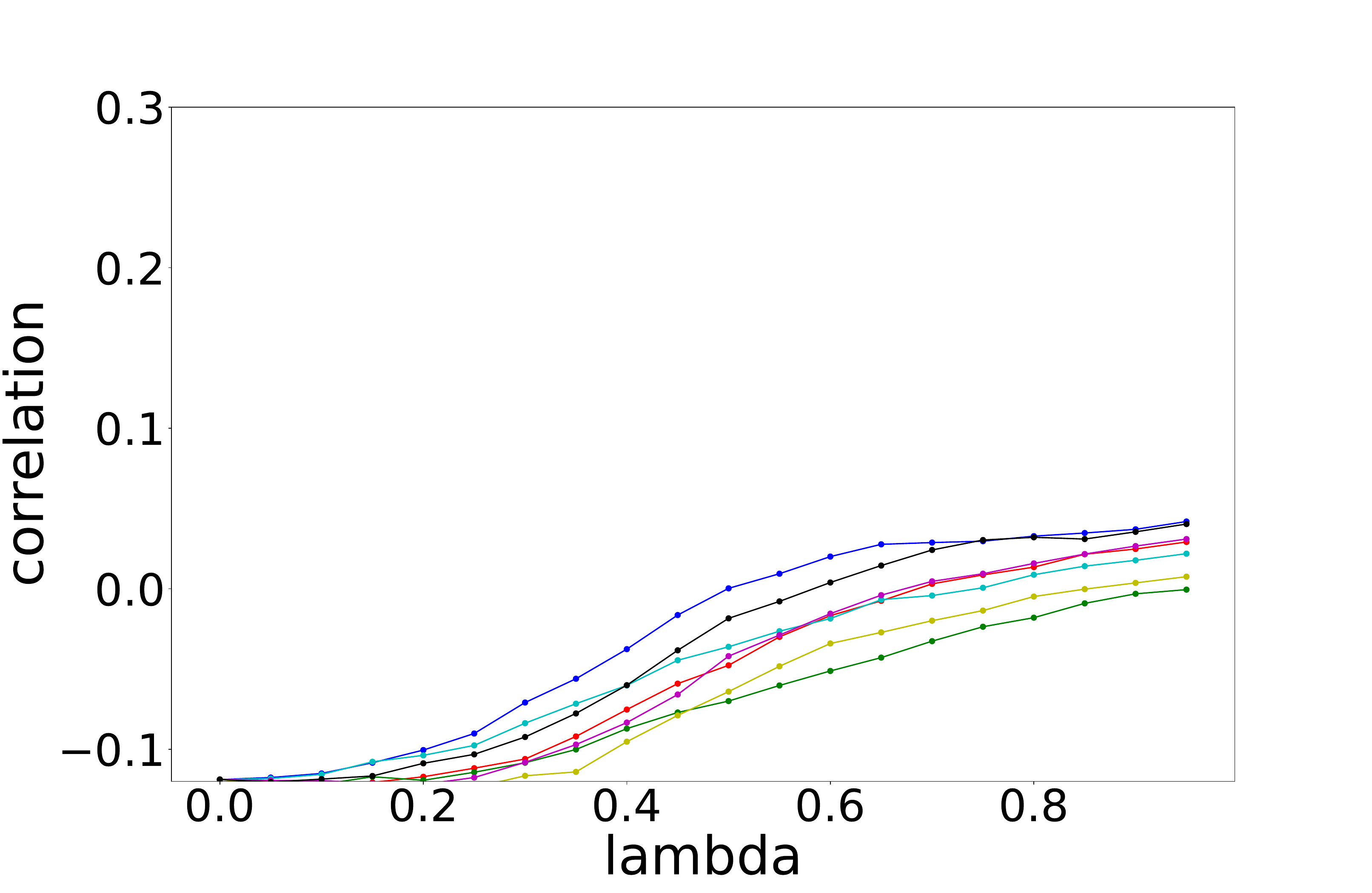}
            \caption{MOR - Fluency}
        \end{subfigure}
        \hfill%
        \begin{subfigure}[b]{0.32\textwidth}
              \includegraphics[width=\linewidth,trim={10 10 155 105},clip]{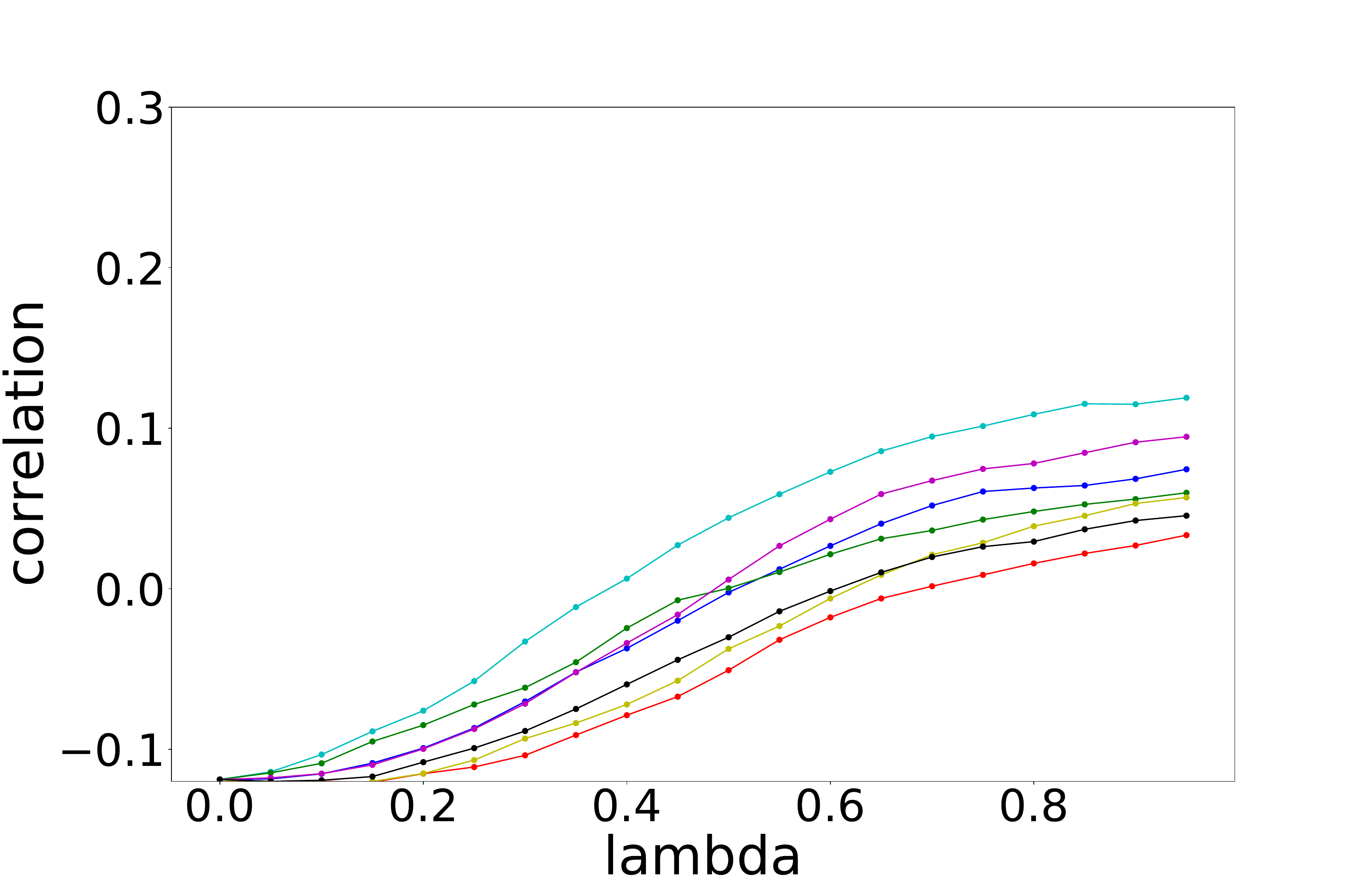}
              \caption{HOR - Fluency}
        \end{subfigure}
        \vspace{-1.5 mm}
        \caption{Kendall Tau ($\tau$) correlation coefficient when lambda ($\lambda)$ $\in [0, 1]$ from (a)-(c) for consistency, (d)-(f) for relevance, (g)-(i) for coherence and (j)-(l) for fluency dimension when BERTScore is used as redundancy penalty for less overlapping reference (LOR), medium overlapping reference (MOR) and high overlapping reference (HOR).}
        \label{fig:bertscore}
        \vspace{-1.5 mm}
\end{figure*}

\end{document}